\documentclass[lettersize,journal]{IEEEtran}

\usepackage{orcidlink}
\usepackage{hyperref}
\hypersetup{hidelinks}
\usepackage{hyperref}
\usepackage{amsmath,amsfonts}
\usepackage{diagbox}
\usepackage[capitalize,noabbrev]{cleveref}
\usepackage{amssymb}

\usepackage{mathtools}
\usepackage{multirow}
\usepackage{float}

\usepackage[figuresright]{rotating}
\usepackage{graphicx}
\usepackage{caption}

\usepackage{multicol}
\usepackage{paralist}
\usepackage{bm}
\usepackage{enumitem}
\usepackage{multirow}
\usepackage{color}
\usepackage{theorem}
\newtheorem{theorem}{Theorem}
\newtheorem{lemma}{Lemma}

\newtheorem{proof}{Proof}
\usepackage{algorithm}

\usepackage{algorithmic}

\usepackage{ifpdf}

\usepackage{booktabs}

\usepackage{array}
\usepackage[caption=false,font=normalsize,labelfont=sf,textfont=sf]{subfig}
\usepackage{textcomp}
\usepackage{stfloats}
\usepackage{url}
\usepackage{verbatim}

\usepackage{bm}

\usepackage{ifpdf}

\usepackage{booktabs}

\newtheorem{proposition}[theorem]{Proposition}

\def\BibTeX{{\rm B\kern-.05em{\sc i\kern-.025em b}\kern-.08em
    T\kern-.1667em\lower.7ex\hbox{E}\kern-.125emX}}
\usepackage{balance}

\begin{document}

\pdfoutput=1
\ifpdf
\title{Sparse Variational Student-t Processes for Heavy-tailed Modeling}

\author{Jian Xu \orcidlink{0000-0002-0350-5528}, 
Delu Zeng* \orcidlink{0000-0001-7322-1873}, 
John Paisley
\thanks{This is the extended version of the AAAI 2024 paper \cite{xu2024sparsest}.}
\thanks{Jian Xu is with the School of Mathematics, South China University of Technology in Guangdong Province, China (email: 2713091379@qq.com).}
\thanks{Delu Zeng is with the the School of Electronic and Information Engineering,
South China University of Technology, Guangzhou 510006, China, and also presently a visiting scholar at the
Department of Electrical and Computer Engineering, University of Waterloo,
Waterloo, ON N2L3G1, Canada (e-mail: dlzeng@scut.edu.cn).}
\thanks{John Paisley is with the Department of Electrical Engineering, Columbia University, New York, USA (email: jwp2128@columbia.edu).}
\thanks{*Corresponding author: Delu Zeng, dlzeng@scut.edu.cn}.}

\maketitle
\begin{abstract}
The Gaussian process (GP) is a powerful tool for nonparametric modeling, but its sensitivity to outliers limits its applicability to data distributions with heavy-tails. Student-t processes offer a robust alternative for heavy-tail modeling, but they lack the scalable developments of the GP to large datasets necessary for practical applications. We present Sparse Variational Student-t Processes (SVTP), the first principled framework that extends the sparse inducing point method to the Student-t process. We develop two novel inference algorithms, SVTP-UB and SVTP-MC, with theoretical guarantees, and derive a natural gradient optimization that exploits a previously unused connection between the Fisher information matrix of the multivariate Student-t distribution and the beta function (the ``beta link''). Experiments on UCI and Kaggle datasets demonstrate that SVTP significantly outperforms sparse GPs on when the data is contains outliers and heavy tails, achieving up to 3× faster convergence and 40\% lower prediction error while maintaining computational efficiency for datasets with over 200,000 samples.
\end{abstract}


\section{Introduction}
Gaussian processes (GPs) \cite{rasmussen2003gaussian} are a versatile, nonparametric method for function approximation that have been applied to a variety of areas such as time-series forecasting \cite{heinonen2018learning}, computer vision \cite{blomqvist2020deep}, and robotics \cite{deisenroth2013gaussian, lee2022trust}. However, GPs are inherently sensitive to outliers due to their reliance on the Gaussian distribution. This limits their robustness in settings where heavy-tailed noise and anomalous observations frequently arise, such as in financial data, multimodal information \cite{song2025diffcl}, hyperspectral images \cite{song2025mcfnet,xiao2025spatial}, and  behavior analysis \cite{song2023multi}. This motivated the development of the Student-t process (TP) \cite{shah2014student}, a heavier-tailed generalization of the GP that offers improved robustness against outliers \cite{tang2017student, chen2020multivariate, andrade2023robustness}. Despite their modeling advantages and comparable theoretical complexity to GPs, the practical deployment of TPs has been hindered by the absence of an efficient and scalable sparse framework. This is due to a more complex probability distribution and lack of established formulations for conditional and marginal distributions using the inducing point method familiar to scalable GPs.

To bridge this gap, we introduce the Sparse Variational Student-t Process (SVTP), a model that achieves robustness to outliers and computational efficiency. SVTP extends the sparse variational GP \cite{titsias2009variational, hensman2015scalable, xu2024sparse} to TPs by incorporating inducing points learnable with variational inference. We propose two strategies for evaluating the variational lower bound: SVTP-UB, which applies Jensen’s inequality to upper bound the KL term, and SVTP-MC, which uses Monte Carlo sampling. We provide a theoretical analysis comparing SVTP with sparse variational GPs (SVGP) that shows why SVTP can more effectively handles outlier-corrupted data. To improve efficiency and scalability of learning, we use natural gradients motivated by information geometry \cite{amari1998natural, hensman2013gaussian}. We derive a tractable Fisher information matrix based on a novel \textit{beta link} using its connection to the beta function. Incorporating natural gradients into modern optimizers such as Adam \cite{kingma2014adam} further improves convergence during training.\newline

\noindent The main contributions of this paper are:
\begin{itemize}[leftmargin=*]
\item \textbf{Sparse Student-t process framework:} We propose a principled sparse approximation for Student-t processes using inducing points, reducing complexity from $\mathcal{O}(n^3)$ to $\mathcal{O}(nm^2)$ while maintaining robustness to outliers.
\item \textbf{Inference algorithms with guarantees:} We propose SVTP-UB (upper bound) and SVTP-MC (Monte Carlo) for computing the variational lower bound, and provide a theoretical analysis detailing when SVTP outperforms SVGP.
\item \textbf{Natural gradients via the beta link:} We establish the connection between the Fisher information matrix of multivariate Student-t distributions and the beta function, enabling scalable natural gradient optimization.
\item \textbf{Empirical validation:} Experiments on UCI and Kaggle datasets demonstrate substantial improvements in convergence speed, accuracy, and robustness to outliers.
\end{itemize}

\section{The Gaussian and Student-t Processes}\label{sec.background}
Gaussian processes (GPs) and Student-t processes (TPs) are probabilistic models widely used in machine learning. While GPs provide analytic tractability, TPs offer greater flexibility in handling heavy-tailed data and outliers. In this section, we review both processes and their sparse representations.

\subsection{Gaussian Processes with Sparse Inducing Points}

A Gaussian process $f(\mathbf{x})$ is defined by a mean function $m(\mathbf{x})$ and kernel function $k(\mathbf{x}, \mathbf{x}')$. In this paper, we take $m(\mathbf{x}) = 0$ without loss of generality. For data $\{(\mathbf{x}_i, y_i)\}_{i=1}^n$, a GP models the predictive distribution at test point $\mathbf{x}_*$ as
\begin{eqnarray}
p(f(\mathbf{x}_*) | \mathbf{X}, \mathbf{y}) &\!=\!& \mathcal{N}(\mu_*,\sigma_*^2),\\
\mu_*& \!=\!& \mathbf{k}(\mathbf{x}_*, \mathbf{X})\mathbf{K}_n^{-1}\mathbf{y},\nonumber\\
\sigma_*^2 &\!=\!&  k(\mathbf{x}_*, \mathbf{x}_*) - \mathbf{k}(\mathbf{x}_*, \mathbf{X})\mathbf{K}_n^{-1}\mathbf{k}(\mathbf{X}, \mathbf{x}_*),\nonumber
\end{eqnarray}
where $\mathbf{K}_n$ is the $n \times n$ pairwise kernel matrix. This has $\mathcal{O}(n^3)$ complexity and uses $\mathcal{O}(n^2)$ memory, limiting scalability. Sparse Variational Gaussian Processes (SVGP) \cite{hensman2015scalable} address this by introducing $m$ inducing points $\mathbf{Z} = \{\mathbf{z}_i\}_{i=1}^m$ in the space of $\mathbf{x}$ with corresponding scalar function values $\mathbf{u} = \{u_i\}_{i=1}^m$. Based on these $m$ points, the approximate predictive distribution then becomes
\begin{eqnarray}
p(f(\mathbf{x}_*)|\mathbf{x}_*, \mathbf{Z},\mathbf{u}) &\!=\!& \mathcal{N}(\mu_z,\sigma^2_z),\\
 \mu_z &\!=\!& \mathbf{k}(\mathbf{x}_*,\mathbf{Z})\mathbf{K}_m^{-1}\mathbf{u}, \nonumber\\
 \sigma^2_z &\!=\!&k(\mathbf{x}_*,\mathbf{x}_*)-\mathbf{k}(\mathbf{x}_*,\mathbf{Z})\mathbf{K}_m^{-1}\mathbf{k}(\mathbf{Z},\mathbf{x}_*).\nonumber
\end{eqnarray}
This reduces complexity to $\mathcal{O}(nm^2)$ and $\mathcal{O}(nm)$ in memory, which provides a significant reduction as typically $m \ll n$.

\subsection{Student-t Processes}

The Student-t process is built on the multivariate Student-t distribution \cite{shah2014student, solin2015state}, and provides heavier tails than the GP controlled by a degree of freedom parameter $\nu > 2$. As $\nu \to \infty$, the TP converges to the GP, enabling flexible modeling of various tail behaviors. The multivariate Student-t distribution with $\nu > 2$ degrees of freedom, mean $\boldsymbol{\mu}\in\mathbb{R}^n$, and scaled covariance $\mathbf{R}\in\mathbb{S}^n_{++}$ has the density function
\begin{equation}
\begin{aligned}
\label{1}
&\hspace{-5pt}\mathcal{ST}(\mathbf{y} |\nu, \boldsymbol{\mu}, \mathbf{R}) =\\&\textstyle ~~~~~~~~~~~\frac{\Gamma\left(\frac{\nu+n}{2}\right) |\mathbf{R}|^{-\frac{1}{2}}}{\Gamma\left(\frac{\nu}{2}\right) \left((\nu-2) \pi\right)^\frac{n}{2}} \left(1 + \frac{(\mathbf{y} - \boldsymbol{\mu})^\top \mathbf{R}^{-1} (\mathbf{y} - \boldsymbol{\mu})}{\nu-2}\right)^{-\frac{\nu+n}{2}}.
\end{aligned}
\end{equation}  
A function follows a \textit{Student-t process}, $f(x) \sim \mathcal{T}\mathcal{P}(\nu, \Psi, K)$, if any finite collection of evaluations $(f(x_1),\dots,f(x_n))$ has a joint multivariate Student-t distribution $\mathcal{ST}(\nu, \Psi, K_n)$, where $K_n$ is the pairwise kernel matrix and $\Psi_n$ is the mean vector. As is evident, the TP follows a similar definition as the GP. It also faces similar $\mathcal{O}(n^3)$ complexity. However, unlike GPs, sparse representations for TPs remain underexplored. The conditional predictive distribution of the multivariate Student-t on which this extension can be built is defined as follows:
\begin{lemma}[Conditional t distribution]
\label{lemma1}
For a Student-t vector $y \sim \mathcal{ST}(\nu,\phi,K)$ partitioned into $y_1$ and $y_2$ of dimensions $n_1$ and $n_2$, the conditional distribution of $y_2$ given $y_1$ is
$$y_2|y_1 \sim \mathcal{ST}\Big(\nu+n_1,\phi_2^*,\frac{\nu+\beta_1-2}{\nu+n_1-2}k^*_{22}\Big),$$ 
where
\begin{equation}
\begin{aligned}
\phi_2^* &= K_{21}K_{11}^{-1}(y_1 - \phi_1) + \phi_2, \\
\beta_1 &= (y_1 - \phi_1)^\top K_{11}^{-1}(y_1 - \phi_1), \\
k^*_{22} &= K_{22} - K_{21}K_{11}^{-1}K_{12}. 
\end{aligned}
\end{equation}
\end{lemma}
This conditional form, which is analogous to the Gaussian, enables the development of sparse inducing point representations for the TP, discussed in the next section.

\section{Sparse Student-t Process and Inference}\label{sec.svtp}
We next propose a sparse Student-t process and variational inference algorithm for scalable posterior approximation and prediction. First, we define inducing points for TPs and construct a variational lower bound objective function. We then present gradient-based optimization algorithms for model inference. A brief list of notational symbols is shown in Table \ref{tab:notation} for reference.

\begin{table}[t]
\centering
\resizebox{.9\columnwidth}{!}{
\begin{tabular}{cl}
\hline
\textbf{Symbol} & \textbf{Description} \\
\hline
$\nu$ & Degrees of freedom (prior) \\
$\tilde{\nu}$ & Degrees of freedom (variational) \\
$M$ & Number of inducing points \\
$N$ & Number of training data points \\
$Z$ & Inducing point locations $(z_1,\ldots,z_M)$ \\
$\mathbf{u}$ & Inducing point function values \\
$\mathbf{f}$ & Function values at training inputs \\
$\mathbf{m}$ & Variational mean parameter \\
$\mathbf{S}$ & Variational covariance parameter \\
$\beta$ & Quadratic form $\mathbf{u}^\top K_{Z,Z'}^{-1}\mathbf{u}$ \\
$K_{Z,Z'}$ & Kernel matrix at inducing points \\
$q(\mathbf{u})$ & Variational distribution $\mathcal{ST}(\tilde{\nu}, \mathbf{m}, \mathbf{S})$ \\
$p(\mathbf{u})$ & Prior distribution $\mathcal{TP}(\nu, \mathbf{0}, K_Z)$ \\
\hline
\end{tabular}}
\caption{A list of key notations.}
\label{tab:notation}
\end{table}

\subsection{Defining Inducing Points for the TP}
Following the sparse GP framework, we introduce $M$ inducing points $Z = (z_1,\dots,z_M)$ in the same space as $\textbf{x}$. These have corresponding function values $\mathbf{u} = (u_1,\dots,u_M)$ that follow a zero-mean Student-t process,
\begin{equation}
    \mathbf{u}\sim \mathcal{TP}(\nu,0,K_{Z}).
\end{equation}
Given $N$ training data points $\{(\mathbf{x}_i, y_i)\}_{i=1}^N$, we model $y_i = f(\mathbf{x}_i)+\epsilon_i$ with noise term $\epsilon_i$. Let $\mathbf{f}=\{f_i\}_{i=1}^N$ denote function values at the training inputs. To reduce the $\mathcal{O}(N^3)$ complexity the comes with inverting the full kernel matrix, we define the joint distribution of $\mathbf{f}$ and $\mathbf{u}$ as
\begin{equation}
    p(\mathbf{u},\mathbf{f}) = \mathcal{ST}\left(\nu, \begin{bmatrix} \mathbf{0} \\ \mathbf{0} \end{bmatrix}, \begin{bmatrix} K_{Z,Z'} & K_{Z,X} \\ K_{X,Z} & K_{{X,X'}} \end{bmatrix}\right),
\end{equation}
and then, by Lemma \ref{lemma1}, work with the conditional distribution,
\begin{equation}
\label{condi}
p(\mathbf{f}|\mathbf{u})=\mathcal{ST}\Big(\nu+M, \mu, \frac{\nu+\beta-2}{\nu+M-2}\Sigma\Big),
\end{equation}
where $\mu=K_{X,Z}K_{Z,Z'}^{-1}\mathbf{u}$, $\Sigma=K_{X,X'}-K_{X,Z}K_{Z,Z'}^{-1}K_{Z,X}$ and $\beta=\mathbf{u}^\top K_{Z,Z'}^{-1}\mathbf{u}$. This enables efficient posterior computation by restricting the pairwise kernel to inducing points.

\subsection{Constructing the Evidence Lower Bound}
We use variational inference to approximate the intractable marginal likelihood $\log p(\mathbf{y})$. The posterior over which this marginalization takes place, $p(\mathbf{u},\mathbf{f}|\mathbf{y})$, factorizes as $p(\mathbf{f}|\mathbf{u},\mathbf{y})p(\mathbf{u}|\mathbf{y})$. We approximate the first term with $p(\mathbf{f}|\mathbf{u})$ as done in the sparse GP. To approximate the second term, we introduce the variational distribution $q(\mathbf{u})$ and construct a mean field approximation to the posterior $p(\mathbf{u},\mathbf{f}|\mathbf{y}) \approx p(\mathbf{f}|\mathbf{u})q(\mathbf{u})$. By Jensen's inequality, the variational lower bound $\mathcal{L}(q)$ on the log evidence (ELBO) is
\begin{equation}
\label{elbo}
    \log p(\mathbf{y}) \geq \mathbb{E}_{p(\mathbf{f}|\mathbf{u})q(\mathbf{u})}[\log p(\mathbf{y}|\mathbf{f})] - \mathrm{KL}(q(\mathbf{u})\|p(\mathbf{u})).
\end{equation}

\subsubsection{Optimization of the ELBO}
We choose a Student-t distribution as the variational distribution $q(\mathbf{u})=\mathcal{ST}(\tilde{\nu}, \mathbf{m},\mathbf{S})$, which matches the t-distribution structure of both prior $p(\mathbf{u})$ and conditional $p(\mathbf{f}|\mathbf{u})$. The ELBO decomposes as
\begin{equation}
\label{elbo1}
    \mathcal{L}(q)=\underset{\text{expected log-likelihood}}{\underbrace{\mathbb{E}_{p(\mathbf{f}|\mathbf{u})q(\mathbf{u})}[\log p(\mathbf{y}|\mathbf{f})]}}-\underset{\text{KL regularization}}{\underbrace{\mathrm{KL}(q(\mathbf{u})\|p(\mathbf{u}))}}
\end{equation}
To enable efficient sampling, we use the reparameterization trick \cite{shah2014student,popescu2022matrix}: Sample $\boldsymbol{\epsilon}_0 \sim \mathcal{N}(0,I)$ and $r^{-1} \sim \Gamma(\nu/2, 1/2)$, and then obtain $\mathbf{u} \sim q(\mathbf{u})$ via the transform
\begin{equation}
    \mathbf{u}=(r(\nu-2)\mathbf{S})^{1/2}\boldsymbol{\epsilon}_0+\mathbf{m},
\end{equation}
where the square root denotes Cholesky decomposition. The vector $\mathbf{u}$ has the required $\mathcal{ST}(\tilde{\nu}, \mathbf{m},\mathbf{S})$ distribution. This allows Monte Carlo estimation of the expected log-likelihood. Lemmas \ref{lemmas} and \ref{lemma2} below provide theoretical justification.
\begin{lemma}
\label{lemmas}
Let $K_n$ be an $n \times n$ symmetric positive definite matrix, and $\phi \in \mathbb{R}^n$, $\nu >0$, $\rho >0$. If
\begin{equation}
r^{-1}  \sim \Gamma(\nu / 2, \rho / 2), \quad
\boldsymbol{y} \mid r  \sim \mathcal{N}(\boldsymbol{\phi}, r(\nu-2) K_n / \rho),
\end{equation}
then marginally $
\boldsymbol{y}\sim \mathcal{S} \mathcal{T} (\nu,\phi,K_n)$.   \end{lemma}
\begin{lemma}
\label{lemma2}
Let $A \in \mathbb{R}^{n\times n}$, $b \in \mathbb{R}^n$, $\mu \in \mathbb{R}^n$, $\Sigma \in \mathbb{R}^{n\times n}$. Let $X \in \mathbb{R}^n$ have distribution $\mathcal{N}(\mu,\Sigma)$. Then the random vector $Y = AX + b$ is distributed as $\mathcal{N}(A\mu + b,A^\top \Sigma A)$.
\end{lemma}

The KL regularization term in the ELBO can be seen as an expectation of the log density ratio and written as
\begin{equation}
\label{KL1}
    \mathrm{KL(}q(\mathbf{u})\| p(\mathbf{u}))=\mathbb{E} _{q(\mathbf{u})}[\log q(\mathbf{u})-\log p(\mathbf{u})].
\end{equation}
To approximate Equation (\ref{KL1}), we can use Monte Carlo sampling to obtain an unbiased estimate that will allow us to employ the reparameterization trick to sample from the variational posterior $q(\mathbf{u})$. In this paper, we refer to this option as SVTP-MC. However, since $q(\mathbf{u})$ is a high-dimensional distribution, Monte Carlo sampling may work poorly with a small number of samples. Therefore, we also propose a method useful for regression tasks with relatively smaller training sets. The general idea is to compute an explicit upper bound for Equation (\ref{KL1}) and use it as a regularizer in place of Equation (\ref{KL1}), which further lower bounds the ELBO. To that end, we rewrite Equation (\ref{KL1}) as
\begin{equation}
\label{KL}
\begin{aligned}
	&\hspace{-5pt}\mathbb{E} _{q(\mathbf{u})}[\log q(\mathbf{u})-\log p(\mathbf{u})] =\\
	&~~~\textstyle\underset{C\left( \nu,\tilde{\nu},\mathbf{S} \right)}{\underbrace{\begin{array}{c}
	\frac{1}{2}\log \frac{|K_{Z,Z'}|}{|\mathbf{S}|}+\frac{M}{2}\log \frac{\nu -2}{\tilde{\nu}-2}+\log \frac{\Gamma \left( \frac{\nu}{2} \right)\Gamma \left( \frac{\tilde{\nu}+M}{2} \right)}{\Gamma \left( \frac{\tilde{\nu}}{2} \right) \Gamma \left( \frac{\nu +M}{2} \right)}
   \end{array}}}\\
	&~~~-\tfrac{\tilde{\nu}+M}{2}\underset{\mathcal{L} _1}{\underbrace{\textstyle\mathbb{E} _{q(\mathbf{u})}\left\{ \log \left[ 1+\frac{1}{\tilde{\nu}-2}\left( \mathbf{u}-\mathbf{m} \right) ^{\mathrm{T}}\mathbf{S}^{-1}\left( \mathbf{u}-\mathbf{m} \right) \right] \right\} }}\\
	&~~~+\tfrac{\nu +M}{2}\underset{\mathcal{L} _2}{\underbrace{\textstyle\mathbb{E} _{q(\mathbf{u})}\left\{ \log \left[ 1+\frac{1}{\nu -2}\mathbf{u}^{\mathrm{T}}K_{Z,Z'}^{-1}\mathbf{u} \right] \right\} }}.
\end{aligned}
\end{equation}
We recall that $\tilde{\nu}$ indicates the variational parameter, while $\nu$ is from the prior for $\mathbf{u}$. Equation (\ref{KL}) requires the computation of two expectations, $\mathcal{L}_1$ and $\mathcal{L}_2$, which represent the entropy of $q(\mathbf{u})$ and the cross-entropy between $q(\mathbf{u})$ and $p(\mathbf{u})$, respectively. First, for $\mathcal{L}_1$, we can establish its relationship with  the standard Student-t distribution using the following lemma.
\begin{lemma}
\label{lemma3}
     Let $ p_X(\boldsymbol{x})=|\Sigma|^
{-\frac{1}{2}}p_{X_0}(\Sigma^{-\frac{1}{2}}
(\boldsymbol{x}-\mu))$ be a location-scale probability density function
where $\mu \in \mathbb{R}^n$ is the location vector and $\Sigma \in  \mathbb{R}^ {n\times n}$ is the dispersion matrix. Let $X_0 =
\Sigma^{-\frac{1}{2}}
(X-\mu )$ be a standardized version of $X$ with density function
$p_{X_0} (\boldsymbol{x}_0)$ not dependent on $(\mu,\Sigma )$. Then
\begin{equation}
    \mathbb{E}_{p(X)}[f(\Sigma^{-\frac{1}{2}}
(X-\mu ))]=\mathbb{E}_{p(X_0)}[f(X_0)],
\end{equation}
where $f$  is any arbitrary continuous function.
\end{lemma}

Next, we can use the following lemma to demonstrate that the entropy of the standard Student-t distribution can be expressed in terms of the gamma function with respect to $\nu$.
\begin{lemma}
\label{lemma4}
Let $X_0$ be an $n$-dimensional standard Student-t
random vector, $p(X_0)=\mathcal{ST}(\nu,0,I)$. Then
\begin{equation}\textstyle
  \mathbb{E} _{p\left( X_0 \right)}\left[ \log \left( 1+\frac{{X_0}^{\top}X_0}{\nu-2} \right) \right] =\Psi \left( \frac{\nu +n}{2} \right) -\Psi \left( \frac{\nu}{2} \right) 
\end{equation}
where $\Psi( a ) =\frac{d\log \Gamma \left( a \right)}{dx}$ for $a \in \mathbb{R}$. 
\end{lemma}
Using Lemmas \ref{lemma3} and \ref{lemma4}, we obtain for $\mathcal{L}_1$ that
\begin{equation}
\label{l1}\textstyle
\mathcal{L}_1=\Psi \left( \frac{\Tilde{\nu} +M}{2} \right) -\Psi \left( \frac{\Tilde{\nu}}{2} \right).
\end{equation}
For $\mathcal{L}_2$, since it involves an intractable integral, we use Jensen's inequality to derive an upper bound,
\begin{equation}
    \mathcal{L}_2 \leqslant \log  \mathbb{E} _{q(\mathbf{u})}\left[ 1+\frac{1}{\nu -2}\mathbf{u}^{\mathrm{T}}K_{Z,Z'}^{-1}\mathbf{u} \right].
\end{equation}
This upper bound for $\mathcal{L}_2$, called $\mathcal{L}_2^\star$, can be expressed as
\begin{small}
\begin{equation}
\label{l2}
\mathcal{L}_2^\star =
\log \left\{ 1+\tfrac{1}{\nu-2}\mathrm{Tr}\left( K_{Z,Z'}^{-1}\mathbf{S} \right) +\tfrac{1}{\nu -2}\mathrm{Tr}\left( K_{Z,Z'}^{-1}\mathbf{mm}^\top  \right) \right\}.
\end{equation}
\end{small}

In this paper, we refer to this approach as SVTP-UB because it computes an upper bound for the KL regularization term. Regarding the applicability of these two methods, since the main purpose of the KL regularization term is to prevent overfitting, which often occurs when there is insufficient data, we use SVTP-UB with a larger regularization term in regression tasks with smaller datasets. On the other hand, for regression tasks with larger datasets we use SVTP-MC.

Based on Equations (\ref{l1}) and (\ref{l2}), we obtain a new lower bound on $\log p(\mathbf{y})$, called $\mathcal{L} ^{\star}(q)$, as follows,
\begin{equation}
\begin{aligned}
    \log p(\mathbf{y})\geq ~& \mathbb{E} _{p(\mathbf{f}|\mathbf{u})q(\mathbf{u})}[\log p(\mathbf{y}|\mathbf{f})]-C\left( \nu,\tilde{\nu},\mathbf{S} \right) \\&+\tfrac{\tilde{\nu}+M}{2}\mathcal{L} _1-\tfrac{\nu +M}{2}\mathcal{L} _{2}^{\star}.
\end{aligned}
\end{equation}
By independence, the log-likelihood factorizes into a sum over data points. According to Equation (\ref{condi}), the marginal distribution of $f_i$ is $p(f_i|\mathbf{u})=\mathcal{ST}(\nu+M,\mu_i,\frac{\nu+\beta-2}{\nu+M-2}\Sigma_i)$, where $\mu_i$ and $\Sigma_i$ are only calculated over observation $\textbf{x}_i$. Therefore, we can learn with the entire dataset using stochastic gradient optimization. 

\subsection{ Prediction and Relation to Sparse Variational GPs}

 We present algorithms for posterior sampling and prediction with the Sparse Variational Student-t process (SVTP). Specifically, we first sample $\mathbf{u}$ from $q(\mathbf{u})=\mathcal{ST}(
\widetilde{\nu }, \mathbf{m},\mathbf{S})$, then sample $\mathbf{f}$ from $p(\mathbf{f}|\mathbf{u})$. To make predictions at new input points $X_*$, we compute the mean and covariance matrix of the predictive distribution,
\begin{equation}
    \begin{aligned}
    \mu_*&=\boldsymbol{k}_*^\top K_{Z,Z'}^{-1}\mathbf{u},\\
    \Sigma_*&=\frac{\nu+\beta-2}{\nu+M-2}(k_{**} - \boldsymbol{k}_*^\top K_{Z,Z'} ^{-1}\boldsymbol{k}_*),
    \end{aligned}
\end{equation}
where $\beta=\mathbf{u}^\top  K_{Z,Z'}^{-1}\mathbf{u}$ and $\boldsymbol{k}_*$ denotes the kernel vector between inducing points $Z$ and the new input points $X_*$, while $k_{**}$ is the kernel between the new input points. We can the obtain the predictive distribution $\mathbf{f}^*\sim\mathcal{ST}(\nu+M,\mu_*,\Sigma_*)$.
We also note that, when we set $q(\mathbf{u})=\mathcal{ST}(
\nu +n, \mathbf{m},\mathbf{S})$, we can analytically marginalize $\mathbf{u}$ resulting in
\begin{equation}
    q( \mathbf{f} ) =\int p( \mathbf{f}|\mathbf{u} ) q( \mathbf{u} ) d\mathbf{u}=\mathcal{ST}( \nu,\mu ',\Sigma ' ),
\end{equation}
where the mean vector $\mu'=K_{X,Z}K_{Z,Z'}^{-1}\mathbf{m}$ and covariance matrix $\Sigma'=K_{X,X'}-K_{X,Z}K_{Z,Z'}^{-1}(K_{Z,Z'}-\mathbf{S})K_{Z,Z'}^{-1}K_{Z,X}$. We provide the derivation in the appendix. The mean and covariance matrix of the marginal distribution $q( \mathbf{f})$ are consistent with SVGP. Furthermore, the following theorem describes the relationship between SVGP and SVTP.
\begin{theorem}
  As $\nu\rightarrow\infty$, the posterior distribution of SVTP converges in distribution to the posterior distribution of SVGP.
\end{theorem}
See the appendix for proof. For a Student-t distribution, the parameter $\nu$ controls the degree of tail-heaviness. Smaller $\nu$ values correspond to heavier tails, while larger $\nu$ values make the tails more closely resemble those of a Gaussian distribution. Therefore, it is a natural extension to apply Student-t distributions to sparse scenarios.

\subsection{Outliers in Regression Analysis}
When evaluating the difference between SVTP and SVGP in handling outliers in regression analysis, we 
follow \cite{hensman2015scalable} for GPs in defining the variational lower bound for SVGP to be
\begin{equation}
\mathcal{L}_{\text{SVGP}}=
\mathbb{E} _{q( \mathbf{u} )}\left[ \log  p( \mathbf{y}|\mathbf{u} ) \right] -\mathrm{KL}\left( q( \mathbf{u})\|p( \mathbf{u} ) \right) .
\end{equation}
For the first term,
\begin{equation}
\label{elbogp}
\begin{aligned}
    \mathbb{E} _{q( \mathbf{u} )}\left[ \log  p( \mathbf{y}|\mathbf{u} ) \right] =&-\tfrac{1}{2}\mathbb{E} _{q( \mathbf{u} )}\left[ ( \mathbf{y}-\mu ) ^\top S^{-1}( \mathbf{y}-\mu ) \right] \\&-\tfrac{1}{2}\log |\Sigma |+C,
\end{aligned}    
\end{equation}
where $\mu,\Sigma$ are defined as in Equation (\ref{condi}) and C is a constant independent of $(X, \mathbf{y})$.
The first term of the ELBO for SVTP, obtained from the previous section, can be defined as
\begin{equation}
\label{elbotp}
\begin{aligned}
    \mathbb{E} _{q( \mathbf{u} )}[ \log  p( \mathbf{y}|\mathbf{u} ) ] = &-\tfrac{\nu +n}{2}\mathbb{E} _{q( \mathbf{u} )} \log  \left( 1+\tfrac{\left( \mathbf{y}-\mu \right) ^\top S^{-1}\left( \mathbf{y}-\mu \right)}{\nu -2} \right) \\&-\tfrac{1}{2}\log |\Sigma |+C.
\end{aligned}    
\end{equation}
The difference between Equations (\ref{elbogp}) and (\ref{elbotp}) is
that the term $( \mathbf{y}-\mu ) ^\top S^{-1}( \mathbf{y}-\mu ) 
$ in  (\ref{elbogp}) is the result
of a log transformation of the term  in  (\ref{elbotp}). If there
are  outliers, this term  would be distorted and
the log transformation can reduce that disturbance. Therefore,
the ELBO of SVTP  is more robust to outliers than that of SVGP.

\section{Optimization of the Evidence Lower Bound}
\label{sec.op}
As discussed in the previous section, we use variational inference to address tractability issues by introducing a distributions \( q_\mathbf{\theta}(\mathbf{u}) \) to approximate the true posterior \( p(\mathbf{u}|\mathbf{f}, \mathbf{y}) \). Here, \(\mathbf{\theta} \in \Omega \subseteq \mathbb{R}^P\) represents the parameters of the distribution \( q \), and \( \Omega \) denotes the parameter space. By applying Jensen's inequality, we derive a variational lower bound \( \mathcal{L}(\mathbf{\theta}) \) on the marginal likelihood,
\begin{equation}
\label{5a}
\begin{aligned}
\log p(\mathbf{y}) \geq \mathbb{E}_{p(\mathbf{f}|\mathbf{u})q_\mathbf{\theta}(\mathbf{u})}[\log p(\mathbf{y}|\mathbf{f})] - \mathrm{KL}(q_\mathbf{\theta}(\mathbf{u}) \| p(\mathbf{u})).
\end{aligned}
\end{equation}
We consider Monte Carlo sampling and gradient-based optimization to maximize the variational objective using steps of the form
\begin{equation}
\label{6}
 \mathbf{\theta}_{t+1}=  \mathbf{\theta}_{t} + \lambda_tA(\mathbf{\theta}_t)\nabla_\mathbf{\theta}\mathcal{L}(\mathbf{\theta})|_{\mathbf{\theta}=\mathbf{\theta}_t},
\end{equation}
where \(\lambda_t \in \mathbb{R}\) denotes the step size and \(A(\mathbf{\theta}_t)\) is  a suitably
chosen  \(P \times P\) positive definite matrix, discussed shortly. On large datasets, the above formula can also be applied using a mini-batch of the data for acceleration.

The vanilla SVTP employs the most straightforward method, which is gradient ascent where \(A(\mathbf{\theta}_t) := I\). The step size can be fixed, decaying, or found by a line search during each iteration. ``Adam'' \cite{kingma2014adam} uses a diagonal matrix with elements equal to $(\sqrt{v_i}+\epsilon)^{-1} m_i$, where $m_i$ and $v_i$ are the bias-corrected exponential moving averages of $\left[\nabla_\mathbf{\theta}\mathcal{L}(\mathbf{\theta})|_{\mathbf{\theta}=\mathbf{\theta}_t}\right]_i$ and $\left(\left[\nabla_\mathbf{\theta}\mathcal{L}(\mathbf{\theta})|_{\mathbf{\theta}=\mathbf{\theta}_t}\right]_i\right)^2$, and the subscript $i$ corresponds to the dimension of the vector. 

While gradient descent and its variants like Adam are effective and widely used, they are not necessarily the most efficient methods since they may not fully exploit the underlying geometry of the parameter space, which can lead to slower convergence and suboptimal performance. To address these limitations, we consider information geometry \cite{amari2012differential}, which provides a framework for developing more efficient optimization algorithms by taking into account the curvature and structure of the parameter space. Next, we discuss how to choose the optimal \(A_t\) from this perspective. 

Assume \(\Omega\) is a Riemannian space, where we can induce a Riemannian metric tensor \(G(\mathbf{\theta})\), a positive definite \(P \times P\) matrix defined by \([G(\mathbf{\theta})]_{ij} = g_{ij}(\mathbf{\theta})\), which generally depends on \(\mathbf{\theta}\). For instance, if \(G(\mathbf{\theta})\) is the identity matrix, \(\Omega\) corresponds to a Euclidean space. With \(G(\mathbf{\theta})\), a distance \(|\cdot|_G\) can be defined in this Riemannian space via the inner product \(\langle \cdot, \cdot \rangle_G\),
\begin{equation}
|d \mathbf{\theta}|_G^2 = \langle d \mathbf{\theta}, d \mathbf{\theta} \rangle_G = \sum\nolimits_{i, j} g_{ij}(\mathbf{\theta}) d\theta_i d\theta_j,
\end{equation}
where \(\theta_i\) and \(\theta_j\) denote the \(i\)-th and \(j\)-th components of \(\mathbf{\theta}\), respectively. The steepest descent direction of a function $\mathcal{L}(\mathbf{\theta})$ at $\mathbf{\theta}$ is defined by the vector $d \mathbf{\theta}$ that minimizes $\mathcal{L}(\mathbf{\theta})$ under the constraint $|d \mathbf{\theta}|_G^2=\varepsilon^2$
for a sufficiently small constant $\varepsilon$. We use the following lemma to prove that setting \(A(\mathbf{\theta}_t)\) in Equation (\ref{6}) to \(G(\mathbf{\theta}_t)^{-1}\) ensures that the loss function \(\mathcal{L}(\mathbf{\theta})\) achieves the steepest ascent in a Riemannian space \cite{amari1998natural}.
\begin{lemma} The steepest ascent direction of \(  \mathcal{L}(\mathbf{\theta}) \) in Riemannian space  is $\widetilde{\nabla} \mathcal{L}(\mathbf{\theta}) = G(\mathbf{\theta})^{-1} \nabla \mathcal{L}(\mathbf{\theta}),$ where \( G = (g_{ij}(\mathbf{\theta})) \) and $\nabla  \mathcal{L}(\mathbf{\theta}) = \big[ \frac{\partial\mathcal{L}(\mathbf{\theta})}{\partial \theta_1}  , \ldots, \frac{\partial  \mathcal{L}(\mathbf{\theta})}{\partial \theta_n} \big]^\top.$ 
\end{lemma}
We define $\widetilde{\nabla} L(\mathbf{\theta})=G(\mathbf{\theta})^{-1} \nabla \mathcal{L}(\mathbf{\theta})$ to be the natural gradient of $\mathcal{L}(\mathbf{\theta})$ in the Riemannian space. Thus, $\widetilde{\nabla} \mathcal{L}(\mathbf{\theta})$ represents the steepest ascent direction of $\mathcal{L}(\mathbf{\theta})$, which corresponds to the natural gradient algorithm in Equation (\ref{6}),
\begin{equation}
\label{zi}
\mathbf{\theta}_{t+1} = \mathbf{\theta}_{t} + \lambda_t G(\mathbf{\theta}_t)^{-1} \nabla_\mathbf{\theta} \mathcal{L}(\mathbf{\theta}) |_{\mathbf{\theta} = \mathbf{\theta}_t}.
\end{equation}
When the space is Euclidean, this corresponds to the steepest ascent  algorithm.

\subsection{Natural Gradient Learning for SVTP}
Computing natural gradients for Student-t models has been difficult because the Fisher information matrix lacks a known closed-form solution. This has limited the use of information-geometric optimization techniques in heavy-tailed models, in contrast to the well-understood Gaussian case. Motivated by this challenge, we derive the Fisher information for the multivariate Student-t variational distribution used in SVTP and show that its elements can be expressed compactly in terms of beta functions. This connection, which we call the ``beta link,'' provides the first tractable method for efficient natural gradients for this problem.

\subsubsection{Fisher Information as Riemannian Metric Tensor}
To implement the natural gradient algorithm, classic information geometry shows that the Riemannian structure of the parameter space  can be defined by the Fisher information \cite{rao1992information,amari2012differential}, which serves as a Riemannian metric tensor on the parameter space. Mathematically, the Fisher information \(F(\mathbf{\theta})\) is a matrix whose \((i, j)\)-th element is given by
\begin{eqnarray}
\label{10}
F_{ij}(\theta) &=& \mathbb{E}_{p(\boldsymbol{x}|\theta)} \left[ \frac{\partial \log p(\boldsymbol{x}|\theta)}{\partial \theta_i} \frac{\partial \log p(\boldsymbol{x}|\theta)}{\partial \theta_j} \right], 
\end{eqnarray}
where \(p(\boldsymbol{x}|\theta)\) is the probability density of a random vector \(\boldsymbol{x}\) given \(\theta\). This measures the ``distance'' between different parameter values \cite{amari1998natural}. 
For example, if  we consider the KL divergence between two distributions and take the small perturbation limit in $\delta$,  we obtain
\begin{equation}
\begin{aligned}\nonumber
    \operatorname{KL}(p_x(\theta)\|p_x(\theta+\delta))=&~\tfrac{1}{2} \delta^{\top}\mathbb{E}[ \nabla_{\theta}^2 \log p_x(\theta)] \delta+\mathcal{O}(\|\delta\|^3)\\=&
-\tfrac{1}{2}|\delta|^2_F+\mathcal{O}(\|\delta\|^3).
\end{aligned}
\end{equation}
Therefore, in a sufficiently small neighborhood $\delta$ around $\boldsymbol{x}$, the KL divergence induces a quadratic norm with curvature given by the expected Hessian of the log density, which corresponds to the Fisher information matrix.
Then we use this matrix \(F(\theta)\) to replace \(G(\theta)\) in Equation (\ref{zi}).

\subsubsection{Representation of  Fisher Information in  SVTP}
In SVTP, we first follow \cite{xu2024sparsest} to parameterize $q_\mathbf{\theta}(\mathbf{u})$ as a Student-t distribution, $\theta =\left\{ \widetilde{\nu },\mathbf{m},\mathbf{S} \right\} \in \Omega \subseteq \mathbb{R}^P$ and $q_\mathbf{\theta}(\mathbf{u})=\mathcal{ST}(\widetilde{\nu }, \mathbf{m},\mathbf{S} )$, where $\mathbf{m}=(m_1,...,m_M)^\top$. Unlike the original SVTP model, we parameterize \(\mathbf{S}\) using a diagonal approximation, 
where the diagonal elements are \(\sigma_1^2, \ldots, \sigma_M^2\) and $M$ is the number of inducing points. We use this approximation to simplify the computation of \( F(\theta) \). Then, \(\theta\) can be expressed in vector form as $\theta = ( m_1, \ldots, m_M, \tilde{\nu}, \sigma_1, \ldots, \sigma_M ) \in \mathbb{R}^{2M+1}$.

Next, we calculate the Fisher information matrix for the variational parameters 
$\theta$. According to Equation (\ref{1}), the log likelihood function of the Student-t distribution can be expressed as follows,
\begin{equation}
\label{13}
\begin{aligned}
\log q_{\mathbf{\theta }}(\mathbf{u})=&-\log \Gamma \left( \tfrac{\tilde{\nu}}{2} \right) +\log \Gamma \left( \tfrac{\tilde{\nu}+M}{2} \right)-\tfrac{M}{2}\log \left( \tilde{\nu}-2 \right) \\&-\tfrac{M}{2}\log \pi 
-\sum\nolimits_{i=1}^M{\log \sigma _i}
\\&-\tfrac{\tilde{\nu}+M}{2}\log \left( 1+\tfrac{1}{\tilde{\nu}-2}\sum\nolimits_{i=1}^M{\tfrac{\left( u_i-m_i \right) ^2}{\sigma _{i}^{2}}} \right), 
\end{aligned}    
\end{equation}
where $\mathbf{u}=(u_1,...,u_M)^\top$. Based on Equation (\ref{10}), we directly calculate the Fisher information matrix using $q(\mathbf{u}| \theta)$.
For \(i = 1, \ldots, M\), we calculate the following derivatives of \(\log q(\mathbf{u}| \theta)\) with respect to \(m_i\), \(\tilde{\nu}\), and \(\sigma_i\):
\begin{equation}
\label{18}
\small{
\begin{aligned}
\frac{\partial \log q(\mathbf{u}|\theta _i)}{\partial m_i}=&-\frac{\tilde{\nu}+M}{\left( \tilde{\nu}-2 \right) \sigma _{i}^{2}}\cdot \frac{u_i-m_i}{1+\frac{1}{\tilde{\nu}-2}\sum_{i=1}^M{\frac{\left( u_i-m_i \right) ^2}{\sigma _{i}^{2}}}},\\
\frac{\partial \log q(\mathbf{u}|\theta _i)}{\partial \sigma _i}=&-\frac{1}{\sigma _i}+\frac{\tilde{\nu}+M}{\left( \tilde{\nu}-2 \right) \sigma _{i}^{3}}\cdot \frac{\left( u_i-m_i \right) ^2}{1+\frac{1}{\tilde{\nu}-2}\sum_{i=1}^M{\frac{\left( u_i-m_i \right) ^2}{\sigma _{i}^{2}}}},\\
\frac{\partial \log q(\mathbf{u}|\theta _i)}{\partial \tilde{\nu}}=&-\frac{1}{2}\Psi \left( \frac{\tilde{\nu}}{2} \right) +\frac{1}{2}\Psi \left( \frac{\tilde{\nu}+M}{2} \right) -\frac{M}{2\left( \tilde{\nu}-2 \right)}\\&+\frac{\tilde{\nu}+M}{2\left( \tilde{\nu}-2 \right) ^2}\cdot \frac{\sum_{i=1}^M{\frac{\left( u_i-m_i \right) ^2}{\sigma _{i}^{2}}}}{1+\frac{1}{\tilde{\nu}-2}\sum_{i=1}^M{\frac{\left( u_i-m_i \right) ^2}{\sigma _{i}^{2}}}},
\end{aligned} 
}
\end{equation}
where $\Psi $ is the digamma function. Given that the Fisher information matrix \( F(\theta) \) is \((2M+1) \times (2M+1)\) and contains three groups of parameters \(\mathbf{m}\), \(\mathbf{S}\), and \(\tilde{\nu}\), we  partition \( F(\theta) \) into blocks as follows:
\begin{equation}
F(\theta)= \left( \begin{matrix}
	F^{\mathbf{m}} & F^{\mathbf{m}\tilde{\nu}} & F^{\mathbf{m}\mathbf{S}} \\
	F^{\tilde{\nu}\mathbf{m}} & F^{\tilde{\nu}} & F^{\tilde{\nu}\mathbf{S}} \\
	F^{\mathbf{S}\mathbf{m}} & F^{\mathbf{S}\tilde{\nu}} & F^{\mathbf{S}}
\end{matrix} \right)
\end{equation}
where $F^{\mathbf{m}}\in \mathbb{R}^{M\times M}, F^{\mathbf{m}\tilde{\nu}}\in \mathbb{R}^{1\times M}, F^{\mathbf{m}\mathbf{S}}\in \mathbb{R}^{M\times M}, 
	F^{\tilde{\nu}\mathbf{m}}\in \mathbb{R}^{M\times 1},  F^{\tilde{\nu}}\in \mathbb{R},  F^{\tilde{\nu}\mathbf{S}}\in \mathbb{R}^{M\times 1}, 
	F^{\mathbf{S}\mathbf{m}}\in\mathbb{R}^{M\times M},  F^{\mathbf{S}\tilde{\nu}}\in \mathbb{R}^{1\times M}, F^{\mathbf{S}}\in\mathbb{R}^{M\times M}$, and the superscripts correspond to the set of parameters required for the derivative of \( \log q(\mathbf{u}| \theta_i) \). For example, the \((i, j)\)-th element of the matrix \( F^{\mathbf{m}} \) is
\begin{equation}
\label{20}
F^{\mathbf{m}}_{ij}(\theta) = \mathbb{E}_{q(\mathbf{u}| \theta)} \left[ \frac{\partial \log q(\mathbf{u}| \theta)}{\partial m_i} \frac{\partial \log q(\mathbf{u}| \theta)}{\partial m_j} \right].    
\end{equation}
Other matrix blocks are defined similarly. Additionally, \( F^{\mathbf{m}\tilde{\nu}} = (F^{\tilde{\nu}\mathbf{m}})^\top \), \( F^{\mathbf{m}\mathbf{S}} = (F^{\mathbf{S}\mathbf{m}})^\top \), and \( F^{\tilde{\nu}\mathbf{S}} = (F^{\mathbf{S}\tilde{\nu}})^\top \). We next solve for these blocks. For the matrix \( F^{\mathbf{m}} \), combining Equations (\ref{18}) and (\ref{20}) we have
\begin{eqnarray}
F_{ij}^{\mathbf{m}}(\theta ) &=&\textstyle\frac{\left( \tilde{\nu}+M \right) ^2}{\left( \tilde{\nu}-2 \right) ^2\sigma _{i}^{2}\sigma _{j}^{2}}\cdot \mathbb{E} _{q( \mathbf{u};\theta )}\Bigg[ \frac{\left( u_i-m_i \right) \left( u_j-m_j \right)}{\big( 1+\frac{1}{\tilde{\nu}-2}\sum_{i=1}^M{\frac{\left( u_i-m_i \right) ^2}{\sigma _{i}^{2}}} \big) ^2} \Bigg]\nonumber\\
&=& C_{i, j}(\theta) \int{\varphi_{i,j}(\mathbf{u},\theta)}
du_1du_2\cdots du_M\label{21}
\end{eqnarray}
where
\begin{eqnarray}
C_{i, j}(\theta) &\triangleq& \frac{\Gamma \left( \frac{\tilde{\nu}+M}{2} \right) \left( \tilde{\nu}+M \right) ^2}{\Gamma \left( \frac{\tilde{\nu}}{2} \right) \pi ^{M/2}\left( \tilde{\nu}-2 \right) ^{2+M/2}\prod_{i=1}^M{\sigma _i}}\cdot \frac{1}{\sigma _{i}^{2}\sigma _{j}^{2}},\nonumber\\
\varphi_{i,j}(\mathbf{u},\theta) &\triangleq&
\frac{\left( u_i-m_i \right) \left( u_j-m_j \right)}{ \left( 1+\frac{1}{\tilde{\nu}-2}\sum_{i=1}^M\sigma _i^{-2}\left( u_i-m_i \right) ^2 \right) ^{\frac{\tilde{\nu}+M}{2}+2}}.\nonumber
\end{eqnarray} 
Using the transformation $\xi_i = \frac{u_i - m_i}{\sigma_i (\tilde{\nu} - 2)^{1/2}}$ we have
$$
\varphi_{i,j} (f(\xi),\theta )=\left( \tilde{\nu}-2 \right) \sigma _i\sigma _j\xi _i\xi _j\Big( 1+\sum\nolimits_{i=1}^M{\xi_{i}^{2}} \Big) ^{-\frac{\tilde{\nu}+M}{2}-2}
$$
and $u_i =f_i(\xi)= m_i + \xi_i \sqrt{\tilde{\nu} - 2} \, \sigma_i$, where $\mathbf{\xi }=(\xi_1,...,\xi_M)^\top$. We can then rewrite Equation (\ref{21}) as
$$ 
F_{ij}^{\mathbf{m}}(\theta )=
C_{i, j}(\theta )\left( \tilde{\nu}-2 \right) ^{\frac{M}{2}}\Big(\prod_{i=1}^M{\sigma _i}\Big)\int{\varphi_{i,j} (f(\xi),\theta )}d\xi_{1:M}.
$$
Before presenting the final Fisher information matrix, we highlight that expressing the Fisher information in terms of beta functions has three important consequences: (i) it enables fully analytical computation of the Fisher information, eliminating the need for high-dimensional numerical integration; (ii) it provides new geometric structure in the parameter space of multivariate Student-t distributions; and (iii) it makes natural gradient optimization practical for SVTP at scale. We refer to
this connection between the Fisher information and the beta function as the
\textit{beta link}.

\subsubsection{Fisher Information and the Beta Function} We next connect the Fisher information to the beta function:
\begin{proposition}
\label{prop:beta_link}
Let $\mathbf{x}\sim t_{\tilde{\nu}}\!\left(\mathbf{m},\mathbf{S}\right)$ and $\mathbf{S}=\mathrm{diag}(\sigma_1^2,\ldots,\sigma_M^2)$ 
be an $M$-dimensional multivariate Student's t-distribution with location $\mathbf{m}$, 
degrees of freedom $\tilde{\nu}>2$, and diagonal scale matrix $\mathbf{S}$.  
Then the Fisher Information Matrix $F(\theta)$ for $\theta=(\mathbf{m},\tilde{\nu},\mathbf{S})$
has the block structure
\begin{equation}
F(\theta)=
\begin{pmatrix}
F^{\mathbf{m}} & \mathbf{0} & \mathbf{0} \\
\mathbf{0} & F^{\tilde{\nu}} & F^{\tilde{\nu}\mathbf{S}} \\
\mathbf{0} & (F^{\tilde{\nu}\mathbf{S}})^\top & F^{\mathbf{S}}
\end{pmatrix}
\end{equation}
with the following equalities: The location block is
\begin{equation}
F_{ii}^{\mathbf{m}}(\theta)= 
\frac{1}{M\sigma_i^2}
\frac{(\tilde{\nu}+M)^2}{\tilde{\nu}-2}
\frac{\mathrm{B}\!\left(\frac{M+3}{2},\,\frac{\tilde{\nu}+1}{2}\right)}
     {\mathrm{B}\!\left(\frac{M}{2},\,\frac{\tilde{\nu}}{2}\right)}.
\end{equation}
The degrees of freedom block is
\begin{equation}
\begin{aligned}
F^{\tilde{\nu}}
=&\;
\alpha(\tilde{\nu})^{2}
+\alpha(\tilde{\nu})\frac{\tilde{\nu}+M}{\tilde{\nu}-2}
\frac{\mathrm{B}\!\left(\frac{M+3}{2},\,\frac{\tilde{\nu}-1}{2}\right)}
     {\mathrm{B}\!\left(\frac{M}{2},\,\frac{\tilde{\nu}}{2}\right)}
\\
&\;
+\frac{(\tilde{\nu}+M)^2}{4(\tilde{\nu}-2)^2}
\frac{\mathrm{B}\!\left(\frac{M+5}{2},\,\frac{\tilde{\nu}-1}{2}\right)}
     {\mathrm{B}\!\left(\frac{M}{2},\,\frac{\tilde{\nu}}{2}\right)},
\end{aligned}
\end{equation}
where
\begin{equation}
\alpha(\tilde{\nu})
=-\frac12\Psi\!\left(\frac{\tilde{\nu}}{2}\right)
+\frac12\Psi\!\left(\frac{\tilde{\nu}+M}{2}\right)
-\frac{M}{2(\tilde{\nu}-2)}.
\end{equation}
The diagonal entries are
\begin{equation}
\begin{aligned}
F_{ii}^{\mathbf{S}}
=&\;
\frac{1}{\sigma_i^{2}}
-\frac{2(\tilde{\nu}+M)}{M+2}
\frac{\mathrm{B}\!\left(\frac{M+3}{2},\,\frac{\tilde{\nu}-1}{2}\right)}
     {\mathrm{B}\!\left(\frac{M}{2},\,\frac{\tilde{\nu}}{2}\right)}
\frac{1}{\sigma_i^{2}}
\\
&\;
+\frac{5(\tilde{\nu}+M)^{2}}{(M+4)(M+2)}
\frac{\mathrm{B}\!\left(\frac{M+5}{2},\,\frac{\tilde{\nu}-1}{2}\right)}
     {\mathrm{B}\!\left(\frac{M}{2},\,\frac{\tilde{\nu}}{2}\right)}
\frac{1}{\sigma_i^{2}},
\end{aligned}
\end{equation}
and off diagonal entries are
\begin{equation}
\begin{aligned}
F_{ij}^{\mathbf{S}}
=&\;
\frac{1}{\sigma_i \sigma_j}
-\frac{2(\tilde{\nu}+M)}{M+2}
\frac{\mathrm{B}\!\left(\frac{M+3}{2},\,\frac{\tilde{\nu}-1}{2}\right)}
     {\mathrm{B}\!\left(\frac{M}{2},\,\frac{\tilde{\nu}}{2}\right)}
\frac{1}{\sigma_i \sigma_j}
\\
&\;
+\frac{(\tilde{\nu}+M)^{2}}{(M+4)(M+2)}
\frac{\mathrm{B}\!\left(\frac{M+5}{2},\,\frac{\tilde{\nu}-1}{2}\right)}
     {\mathrm{B}\!\left(\frac{M}{2},\,\frac{\tilde{\nu}}{2}\right)}
\frac{1}{\sigma_i \sigma_j}.
\end{aligned}
\end{equation}
The cross block is
\begin{equation}
\begin{aligned}
F_{i}^{\tilde{\nu}\mathbf{S}}
=&
\left(
\frac{\alpha(\tilde{\nu})(\tilde{\nu}+M)}{M+2}
-\frac{\tilde{\nu}+M}{4(\tilde{\nu}-2)}
\right)
\tfrac{\mathrm{B}\!\left(\frac{M+3}{2},\,\tfrac{\tilde{\nu}-1}{2}\right)}
     {\mathrm{B}\!\left(\frac{M}{2},\,\tfrac{\tilde{\nu}}{2}\right)}
\frac{1}{\sigma_i}
\\
&\hspace{-10pt}
+\frac{(\tilde{\nu}+M)^2}{2(\tilde{\nu}-2)(M+2)}
\frac{\mathrm{B}\!\left(\frac{M+5}{2},\,\frac{\tilde{\nu}-1}{2}\right)}
     {\mathrm{B}\!\left(\frac{M}{2},\,\frac{\tilde{\nu}}{2}\right)}
\frac{1}{\sigma_i}-\frac{\alpha(\tilde{\nu})}{\sigma_i}.
\end{aligned}
\end{equation}
Moreover, $F^{\mathbf{m}\mathbf{S}}=\mathbf{0}$ and $F^{\mathbf{m}\tilde{\nu}}=\mathbf{0}$. This establishes the closed-form Fisher Information matrix of the multivariate t-distribution and its ``beta link'' representation.
\end{proposition}

Although beta functions are prevalent in multivariate statistics, their
connection to the Fisher information of multivariate Student-t distributions
does not appear to have been previously made. This ``beta link'' therefore provides a novel bridge between classical multivariate statistical theory and modern information geometry.

When calculating natural gradients, we need to invert \(F(\theta)\). This is a block-diagonal matrix with one of its diagonal blocks \(F^{\mathbf{m}}\) a diagonal matrix. Its inversion also follows the same block pattern,
\begin{equation}
 \label{42}
 F(\theta )^{-1}=\left( \begin{matrix}
 	\left( F^{\mathbf{m}} \right) ^{-1}&		O\\
 	O&		\varLambda _{\tilde{\nu}\mathbf{S}}^{-1}\\
 \end{matrix} \right) ,
 \end{equation}
  reducing the inversion to an \((M+1) \times (M+1)\) matrix.

 \subsubsection{Stochastic Natural Gradient}
As shown for stochastic variational inference \cite{hoffman2013stochastic}, in natural gradient ascent we can learn from the entire dataset using stochastic gradients as
 \begin{equation}
 \label{43}
 \mathbf{\theta}_{t+1} = \mathbf{\theta}_{t} - \lambda_t\tfrac{B}{N} F(\mathbf{\theta}_t)^{-1} \nabla_\mathbf{\theta} \mathcal{L}_B(\mathbf{\theta}) |_{\mathbf{\theta} = \mathbf{\theta}_t},    
 \end{equation}
 where \(\mathcal{L}_B(\mathbf{\theta})\) corresponds to the loss function of the mini-batch, with \(B\) denoting the batch size. In addition to the variational parameters \(\theta\), we optimize other model hyperparameters such as kernel hyperparameters jointly using the ELBO. For these parameters, we use Adam to optimize the hyperparameters in conjunction with natural gradients for the variational parameters \(\theta\). We outline our algorithm in Algorithm \ref{algorithm}. By using stochastic natural gradients, the algorithm adapts to the local curvature of the parameter space, improving convergence speed and local optimal solutions.

\begin{algorithm}[t]
\caption{ Stochastic natural gradient for
 SVTP }
\label{algorithm}
\begin{algorithmic}
   \STATE {\bfseries Input:} Training data $X, \mathbf{y}$ of size $N$,
   mini-batch size $B$.
   \STATE {\bfseries Initialize:} Variational parameters $\mathbf{\theta}=\{\tilde{\nu}, \mathbf{u}, \mathbf{S}\}$, inducing points $Z$,  kernel parameter $\eta$.
      \REPEAT
      \STATE Calculate $F(\theta)^{-1}$ of Equation (\ref{42})
   \STATE Sample mini-batch $I \subset\{1, \ldots, N\}$ with $|I|=B$ 
    \STATE Compute $\mathcal{L}_B(\theta,Z,\eta)$ by Equation (\ref{elbo1})
   \STATE Do one stochastic optimization step on $Z,\eta$
    \STATE Do one stochastic natural gradient step on  $\theta$
    \UNTIL{$\theta,Z,\eta$  converge}
    \end{algorithmic}
\end{algorithm}

\subsubsection{Discussion on Complexity}
Since we use a diagonal covariance approximation for the variational parameter \(\mathbf{S}\), the resulting computational complexity is \(\mathcal{O}(M^3 + NM^2)\). Given that the original complexity of the SVTP model is \(\mathcal{O}(NM^2)\), and considering the typical regime where \(N \gg M\), our approach remains computationally efficient. Moreover, stochastic natural gradients are advantageous for large-scale datasets. By leveraging the geometry-aware updates of natural gradients aligned with the direction of steepest ascent in the information geometry, we enhance both convergence speed and scalability of the model.

\section{Related Works}

\paragraph{Stochastic Process Methods for Regression}
Stochastic processes provide a principled mathematical foundation for modeling distributions over functions. Among them, Gaussian Processes (GPs) \cite{rasmussen2003gaussian} have been widely adopted for regression tasks due to their flexibility, nonparametric formulation, and ability to quantify uncertainty in a Bayesian manner. GPs have found successful applications in time-series forecasting \cite{heinonen2018learning,li2025generative}, robotics \cite{deisenroth2013gaussian}, computer vision \cite{blomqvist2020deep,li2025evodiff, xu2024sparse}, density estimation \cite{paisley2025,chen2025dequantified}, and explainable neural networks \cite{zhang2024a}. However, standard GPs rely on the assumption of Gaussian noise and light-tailed distributions, making them sensitive to outliers and poorly suited for modeling heavy-tailed or non-Gaussian data distributions commonly encountered in real-world scenarios.

To address these limitations, a variety of extensions to GPs have been proposed. One such extension is the Student-t Process (TP) \cite{shah2014student}, which generalizes GPs by replacing the Gaussian marginal distributions with Student-t distributions, thereby introducing heavier tails and enhanced robustness to outliers. Several theoretical and empirical studies have demonstrated the advantages of TPs in handling noisy or corrupted data \cite{tang2017student, chen2020multivariate, andrade2023robustness, xu2024multivariate,HaoWang2025GeneralImpute,du2022flow,lin2025reciprocalla}. Another promising direction involves elliptical processes \cite{baankestad2020elliptical, baankestad2023variational}, which generalize Gaussian and Student-t processes by allowing elliptical contours of constant density in the function space. These processes offer a unifying probabilistic framework with greater flexibility in capturing dependencies and tail behaviors.

\paragraph{Natural Gradients}
Gradient-based optimization methods such as stochastic gradient descent (SGD) and Adam \cite{kingma2014adam} are commonly used in variational inference, but they often suffer from slow convergence due to their ignorance of the underlying geometry of the parameter space. To overcome this, Amari \cite{amari1998natural} proposed the natural gradient method, which preconditions the gradient using the inverse Fisher information matrix, yielding optimization paths aligned with the geometry of the variational family. Natural gradients have been successfully applied in variational inference for GPs \cite{hensman2013gaussian, salimbeni2018natural}, optimal transport \cite{li2018natural}, deep generative models \cite{pascanu2013revisiting, shen2020sinkhorn}, and more recently in pre-trained models \cite{wu2024improved} such as vision Transformer \cite{dosovitskiy2020image, liu2021swin} and Lora \cite{hu2022lora}. These approaches have shown faster convergence and improved stability during training. However, to the best of our knowledge, natural gradients have not yet been systematically studied or applied in the context of Student-t processes.

\paragraph{Sparse Bayesian Methods}
Scalability is a major bottleneck in Bayesian models, particularly in stochastic process methods like GPs and TPs, due to their cubic computational cost in the number of training points. Sparse variational methods \cite{titsias2009variational} address this challenge by introducing a set of inducing variables that act as a compact summary of the full dataset. These inducing points enable scalable inference by reducing complexity to linear in the number of training samples, while still capturing the key features of the posterior distribution. This idea has been extended to various models, including Bayesian Neural Networks (BNNs) \cite{hernandez2015probabilistic, ritter2021sparse, jospin2022hands, dabiran2023sparse,xu2024variational,xu2025fully} and deep kernel learning \cite{wilson2016deep,xu2025bayesian}. Recent work has also explored inducing points in more abstract feature spaces \cite{moss2023inducing, xu2024neural,10634990,cheng2025empowering} and connections to random Fourier features \cite{cutajar2017random,paisley2022,zhang2024a,zhang2024b,paisley2025}. Despite these advances, sparse approximations for Student-t processes remain limited, primarily due to the difficulty of computing marginal and conditional distributions. Sparse methods have also proven effective beyond stochastic processes. For example, Song et al. \cite{song2022multimodal} show that sparse attention in multimodal transformer architectures can selectively focus on the most relevant features, highlighting the broader utility of sparsity in scalable high-dimensional models.

\renewcommand{\arraystretch}{1.1}{
\begin{table}[t]

\centering
\resizebox{1\columnwidth}{!}{
\begin{tabular}{c|ccccc}

\hline
Dataset & $n$ & $d$ & $m=n/10$&$m=n/4$  & Full TP\\
\hline
Yacht & 307&6 & 0.042s& 0.045s& 0.058s \\
\hline
Boston & 505&13 & 0.042s& 0.045s& 0.062s \\
\hline
Energy &767&8  & 0.043s& 0.046s& 0.075s \\
\hline
Concrete & 1,029&8 & 0.050s& 0.061s& 0.110s \\
\hline
Kin8nm & 8,192&8 &0.45s & 0.98s& 5.49s \\
\hline
Elevator & 16,599&18 & 0.931s& 2.412s& 97.98s \\
\hline
Protein & 45,730&9 & 12.44s& 97.93s& --- \\
\hline
Taxi & 209,673 &7  & ---& ---& --- \\
\hline
\end{tabular}}

\caption{ Comparison of computational complexity on regression experiments. In this context, $m$ refers to the number of inducing points. ``---" stands for not having enough space and computational power to calculate its time.}
\label{time1}
\end{table}
}

\section{Experiments}
\label{sec:exp}

\begin{figure*}[htbp]
\label{figure1}
  \centering
  \includegraphics[width=\textwidth]{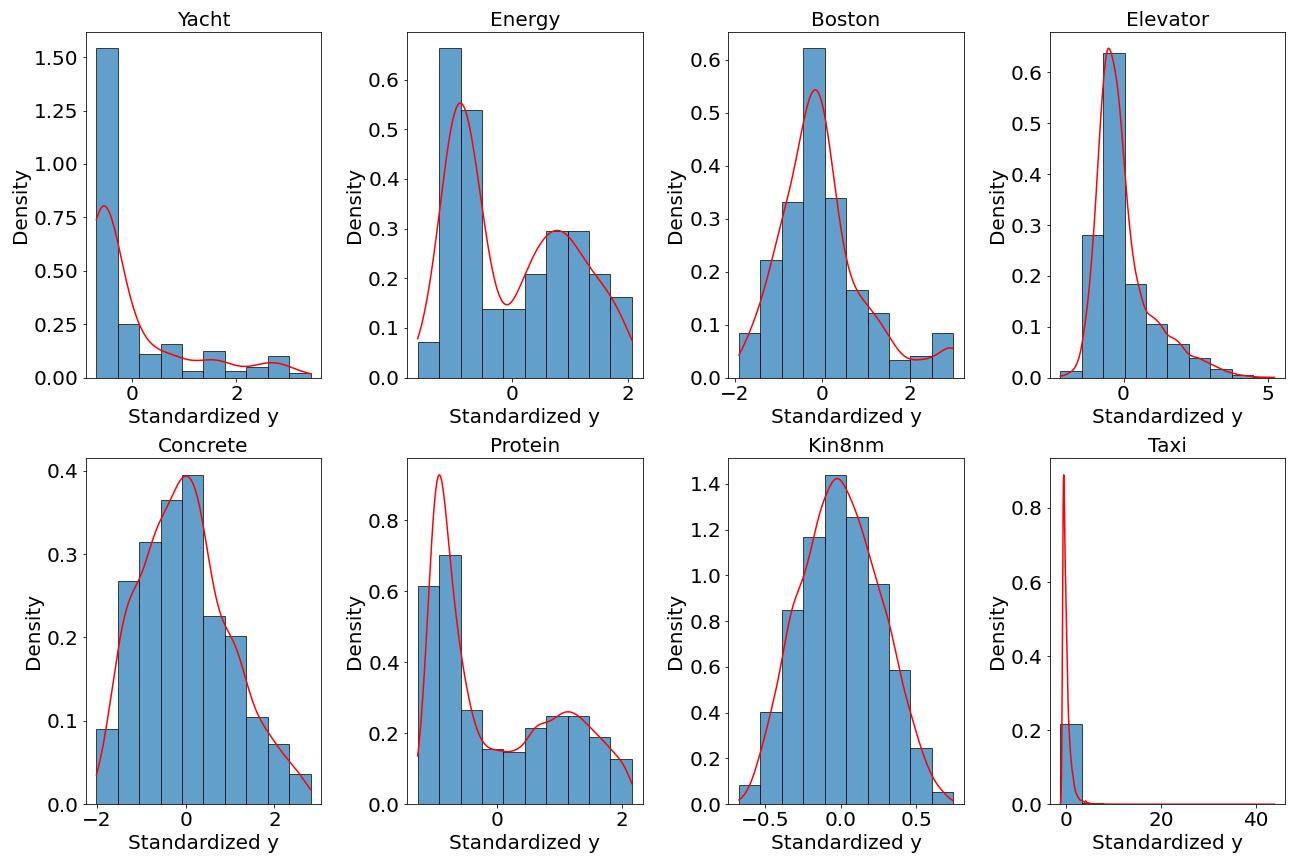}
  \caption{Empirical density analysis and kernel density estimation on the datasets considered.}
  \label{fig:image}
\end{figure*}

We evaluate SVTP-UB and SVTP-MC on eight datasets from UCI and Kaggle. We use an 80/20 train-test split and compare against Sparse Variational Gaussian Processes (SVGP), full Student-t Processes (TP) and SVGP with student-t likelihood (SVGP+T) \cite{martinez2017robust}. For optimization, we compare our Stochastic Natural Gradient Descent (SNGD) against SGD, Adam~\cite{kingma2014adam}, Adagrad~\cite{ward2020adagrad}, Adamax~\cite{kingma2014adam}, and Nadam~\cite{dozat2016incorporating}. All experiments use batch size 1024, learning rate 0.01, standardized data, 5000 maximum iterations, and fixed noise term $\epsilon_i = 0.10$. We implement all methods in PyTorch on a single NVIDIA A100 GPU. Table~\ref{time1} provides dataset sizes, along with running times. Brief descriptions also follow:
\begin{itemize}[leftmargin=.65in]
    \item[\textbf{Concrete}:] Slump flow measurements of concrete mixtures.
    \item[\textbf{Boston}:] Housing prices from Boston suburbs (1978).
    \item[\textbf{Kin8nm}:] Forward kinematics of 8-link robot arm --- highly non-linear.
    \item[\textbf{Yacht}:] Hydrodynamic performance of sailing yachts.
    \item[\textbf{Energy}:] Energy analysis of 768 building shapes with varying parameters.
    \item[\textbf{Elevator}:] F16 aircraft elevator control data.
    \item[\textbf{Protein}:] Physicochemical properties of protein structures (CASP 5-9).
    \item[\textbf{Taxi}:] NYC yellow taxi trip fares prediction.
\end{itemize}

\subsection{Computational Complexity of SVTP and Full TP}
To empirically validate the computational efficiency of SVTP, we conducted experiments on eight benchmark datasets and compared the average time per epoch of SVTP-UB with that of the conventional full TP method. As shown in Table~\ref{time1}, the computational cost decreases substantially as the number of inducing points in SVTP is reduced. Correspondingly, the average epoch time also declines, demonstrating the improved computational efficiency of SVTP.

\renewcommand{\arraystretch}{1.2}{
\begin{table*}[h]
\resizebox{1\textwidth}{!}{
\begin{tabular*}{\linewidth}{l|cccc|cccc}
\hline
\multirow{2}{4em}{Dataset} & \multicolumn{4}{c|}{Predictive Mean Squared Error} & \multicolumn{4}{c}{Test Log Likelihood} \\
& SVGP & SVGP+T & SVTP+UB & SVTP+MC 
& SVGP & SVGP+T & SVTP+UB & SVTP+MC \\ \hline

Yacht & $7.56\pm0.23$ & $2.35\pm0.11$ & $\mathbf{1.44}\pm0.02$ & $1.55\pm0.03$
      & $-485\pm2.4$ & $-165\pm1.6$ & $\mathbf{-114}\pm0.5$ & $-122\pm0.7$ \\ \hline

Energy & $2.02\pm0.06$ & $1.12\pm0.05$ & $\mathbf{0.74}\pm0.04$ & $0.84\pm0.05$
       & $-156\pm1.8$ & $-92.4\pm1.4\,$ & $\mathbf{-63.3\pm1.2}$ & $-70.5\pm1.3$ \\ \hline

Boston & $3.17\pm0.21$ & $2.46\pm0.14$ & $\mathbf{1.82\pm0.10}$ & $2.01\pm0.12$
       & $-171\pm4.8$ & $-148\pm3.1$ & $\mathbf{-126\pm2.6}$ & $-140\pm2.9$ \\ \hline

Elevator & $6.54\pm0.28$ & $2.85\pm0.17$ & $1.93\pm0.15$ & $\mathbf{1.84\pm0.11}$
         & $-391\pm19.2$ & $-195\pm9.8$ & $-126.1\pm10.2$ & $\mathbf{-106\pm8.7}$ \\ \hline

Concrete & $7.48\pm0.24$ & $6.12\pm0.19$ & $5.71\pm0.20$ & $\mathbf{3.23\pm0.18}$
         & $-478\pm11.4$ & $-412\pm8.6$ & $-435\pm9.0$ & $\mathbf{-276\pm7.4}$ \\ \hline

Protein & $7.23\pm0.26$ & $6.12\pm0.20$ & $5.64\pm0.22$ & $\mathbf{4.86\pm0.14}$
        & $-425\pm19.3$ & $-308\pm13.7$ & $-295\pm15.5$ & $\mathbf{-260\pm9.8}$ \\ \hline

Kin8nm & $5.19\pm0.52$ & $4.65\pm0.49$ & $4.35\pm0.47$ & $\mathbf{3.85\pm0.46}$
       & $-400\pm28.6$ & $-352\pm21.2$ & $-385\pm22.5$ & $\mathbf{-225\pm19}$ \\ \hline

Taxi & $0.76\pm0.07$ & $0.62\pm0.05$ & $0.54\pm0.05$ & $\mathbf{0.41\pm0.04}$
     & $-301\pm8.9$ & $-168\pm6.1$ & $-152\pm6.4$ & $\mathbf{-124\pm4.6}$ \\ \hline

Concrete$^*$ 
& $12.20\pm0.32$ & $9.41\pm0.28$ & $8.62\pm0.25$ & $\mathbf{6.40\pm0.22}$
& $-658\pm21$ & $-412\pm15.9$ & $-471.0\pm14.2$ & $\mathbf{-336\pm14}$ \\ \hline

Kin8nm$^*$ 
& $14.66\pm2.25$ & $10.83\pm1.11~\,$ & $9.12\pm1.72$ & $\mathbf{8.86\pm0.65}$
& $-1169\pm62$ & $-571\pm43.2$ & $-622\pm52.4$ & $\mathbf{-501\pm31}$ \\ \hline

\end{tabular*}
}
\caption{The Predictive Mean Squared Error and Test Log Likelihood performance measures. Concrete$^*$ and Kin8nm$^*$ are selected for inclusion of outliers as described in the text.}
\label{mse}
\end{table*}
}

\subsection{Real-world Regression Problems}
Unlike the baseline approach of \cite{shah2014student}, our method significantly reduces computational complexity, enabling evaluation on entire datasets rather than small subsets. Each experiment was repeated using five-fold cross-validation, and we reported both the average and range of results. Performance was assessed using Mean Squared Error (MSE) and test Log Likelihood (LL). The number of inducing points was set to one-fourth of the dataset size, and a squared exponential kernel was employed. For large datasets such as Protein and Taxi, the inducing points were chosen as one-fourth of the batch size. As presented in Table~\ref{mse}, our SVTP methods consistently outperform SVGP across all datasets, highlighting the improved predictive accuracy and uncertainty quantification of SVTP.

\subsection{Regression Experimental Analysis}
To further analyze the performance gains of SVTP, we compute an empirical density analysis and kernel density estimation on the standardized targets across datasets. As shown in Figure~1, many datasets exhibit varying degrees of outliers, irregularities, and heavy-tailed behavior. Yacht and Taxi show particularly prominent deviations, and SVTP achieves substantial improvements over SVGP on these datasets, reinforcing the robustness of our approach. 

We additionally selected the Concrete and Kin8nm datasets for outlier regression experiments. We synthetically introduced stronger outliers by adding three standard deviations to the targets of 5\% of the data. The results, also reported in Table~\ref{mse}, further demonstrate the robustness of SVTP in the presence of extreme noise. We also conducted an empirical comparison between SVTP-UB and SVTP-MC in computing the KL regularization term. Four datasets---Boston, Elevator, Concrete, and Protein---were used to obtain the converged KL values for both methods, summarized in Table~\ref{KLtable}. SVTP-UB tends to yield larger KL estimates, which, when combined with the results of Table~\ref{mse}, suggests that its tighter regularization hinders convergence on large datasets. In contrast, for smaller datasets where overfitting is of primary concern, these stronger constraints help. Therefore, we recommend using SVTP-MC for large datasets to achieve smoother convergence and SVTP-UB for small datasets to better control overfitting.

\renewcommand{\arraystretch}{.7}{
\begin{table}[b]
\centering
\resizebox{.8 \columnwidth}{!}{
\begin{tabular}{c|c|c}

\hline

{\tiny Datasets} &  {\tiny SVGP+UB} & {\tiny SVGP+MC}\\

\hline
{\tiny Boston} &  {\tiny 56} &{\tiny 2.2} \\
{\tiny Elevator} &   {\tiny 42}& {\tiny 3.1}\\
{\tiny Concrete} &   {\tiny 78}& {\tiny 1.3} \\
{\tiny Protein} &   {\tiny 39}& {\tiny 6.3} \\
\hline

\end{tabular}
}
\caption{A comparison of the average values of the KL divergence term on several datasets.}\label{KLtable}
\end{table}
}

\begin{figure*}[t]
    \centering
        \includegraphics[width=.49\textwidth]{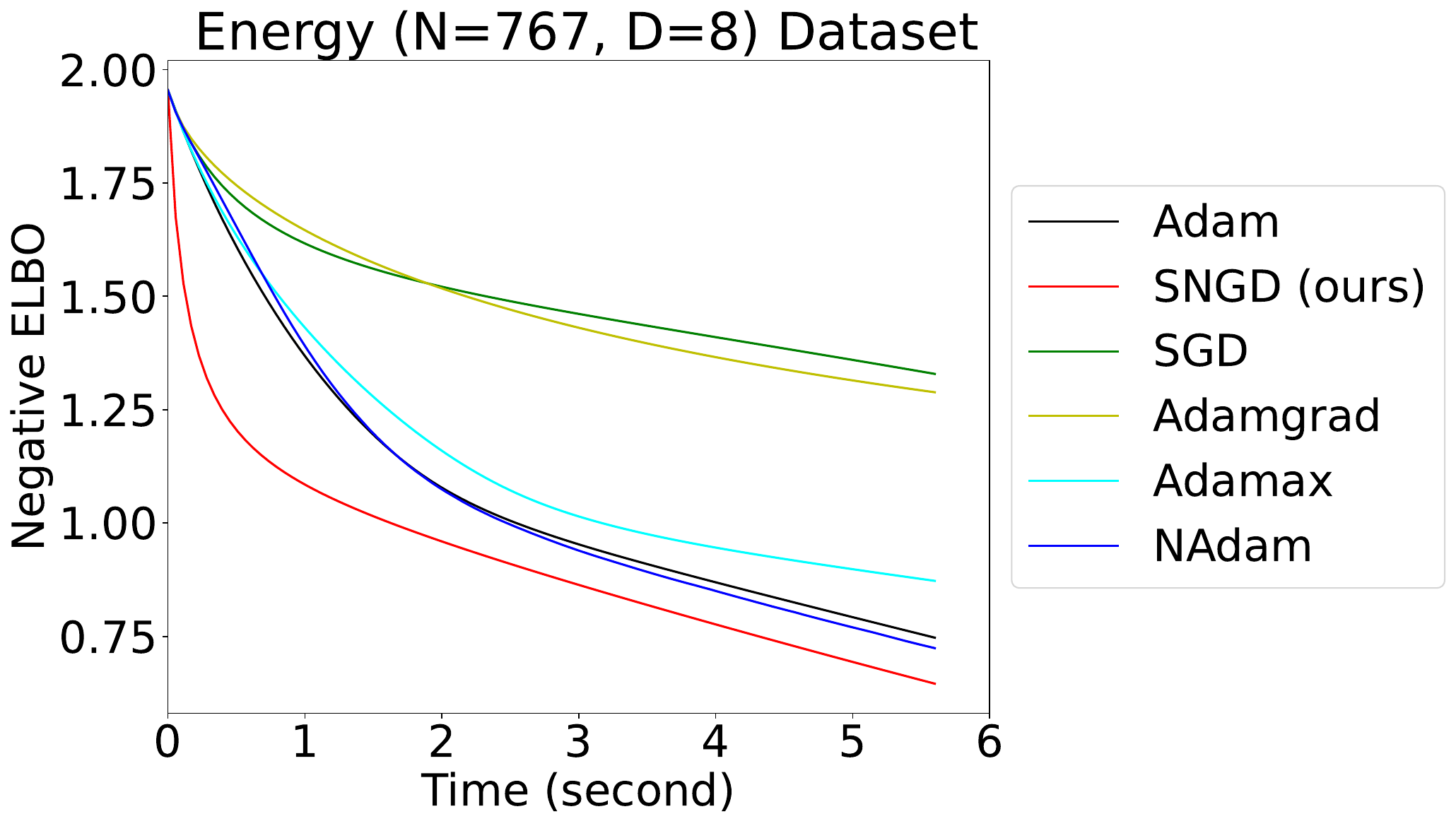
        }\quad
        \includegraphics[width=.49\textwidth]{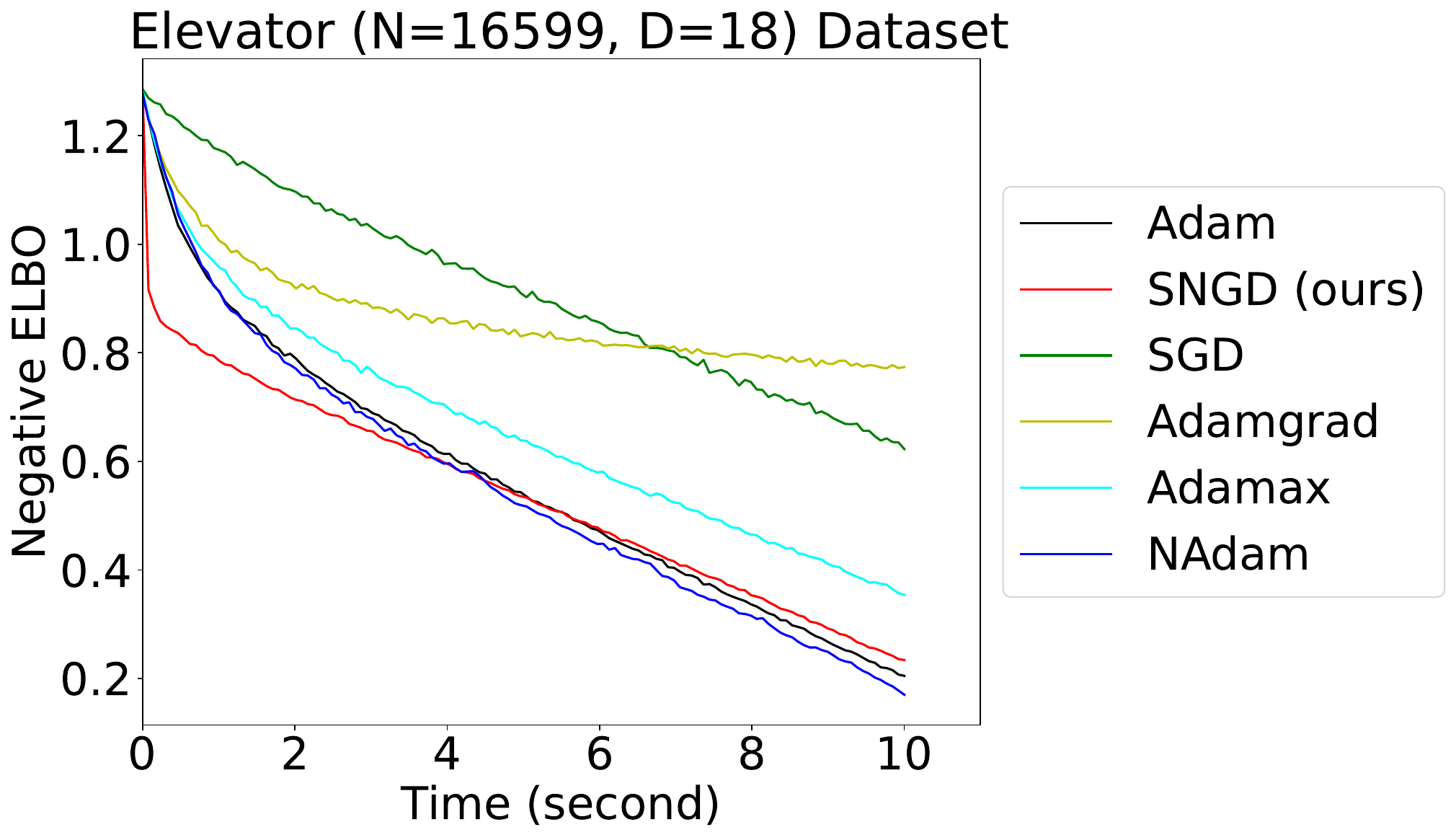}
        \includegraphics[width=.49\textwidth]{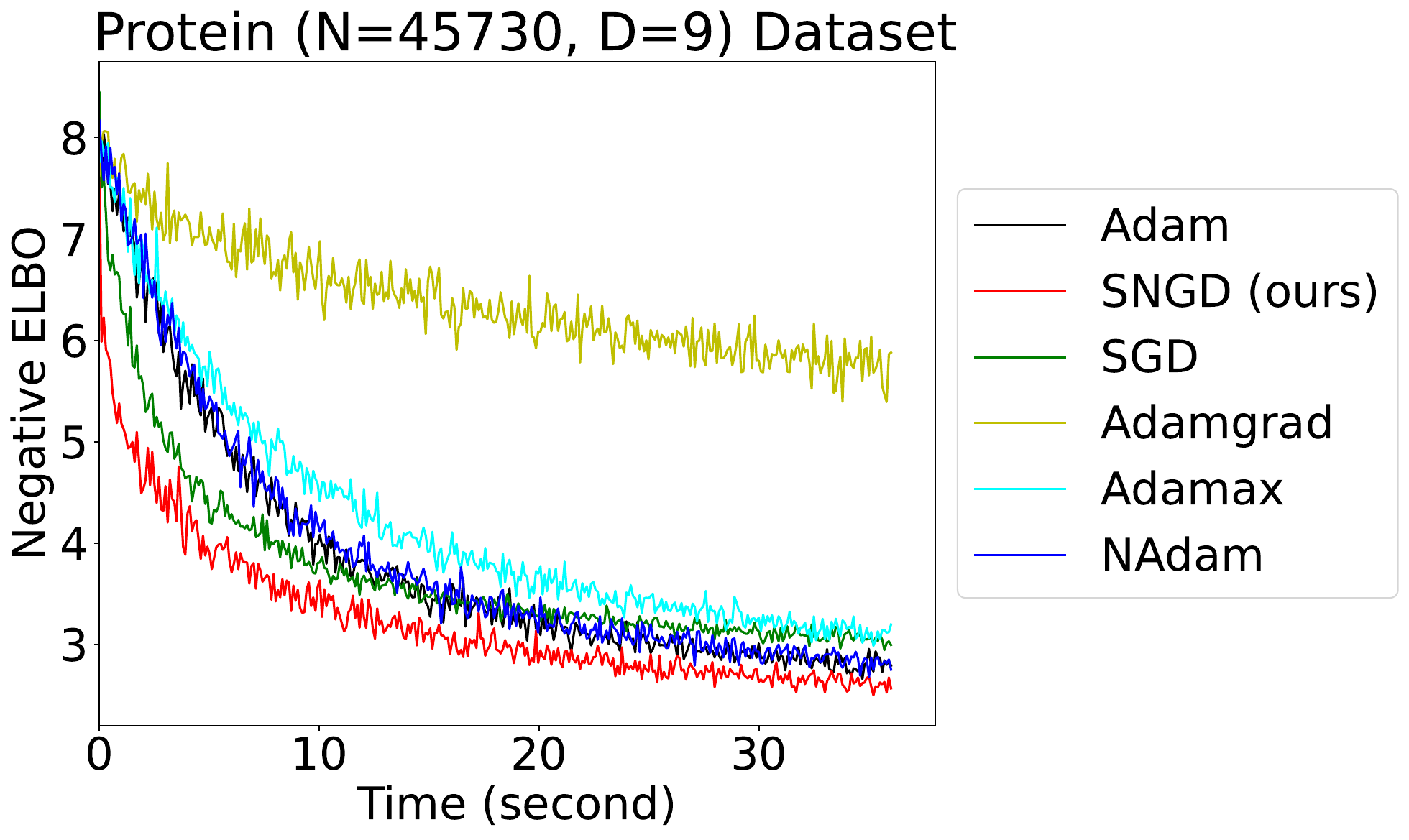}\quad
        \includegraphics[width=.49\textwidth]{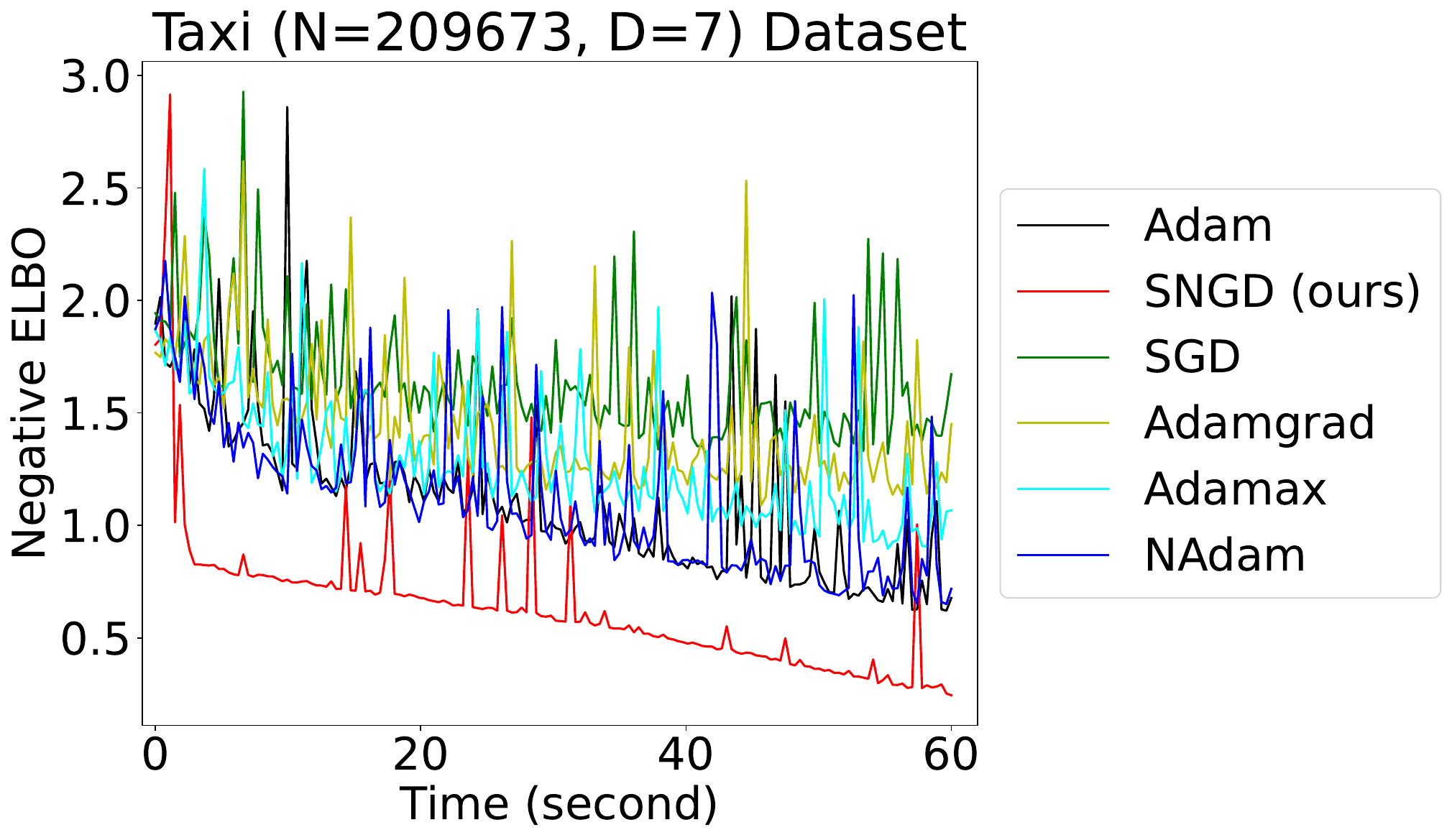}
\caption{Negative ELBO curves for four datasets. Our SNGD method (red) achieves faster convergence compared to five baseline optimizers (Adam, SGD, Adagrad, Adamax, and Nadam), demonstrating improved optimization efficiency.}
\label{fig:main}
\end{figure*}
\begin{figure*}[t]
        \includegraphics[width=.49\textwidth]{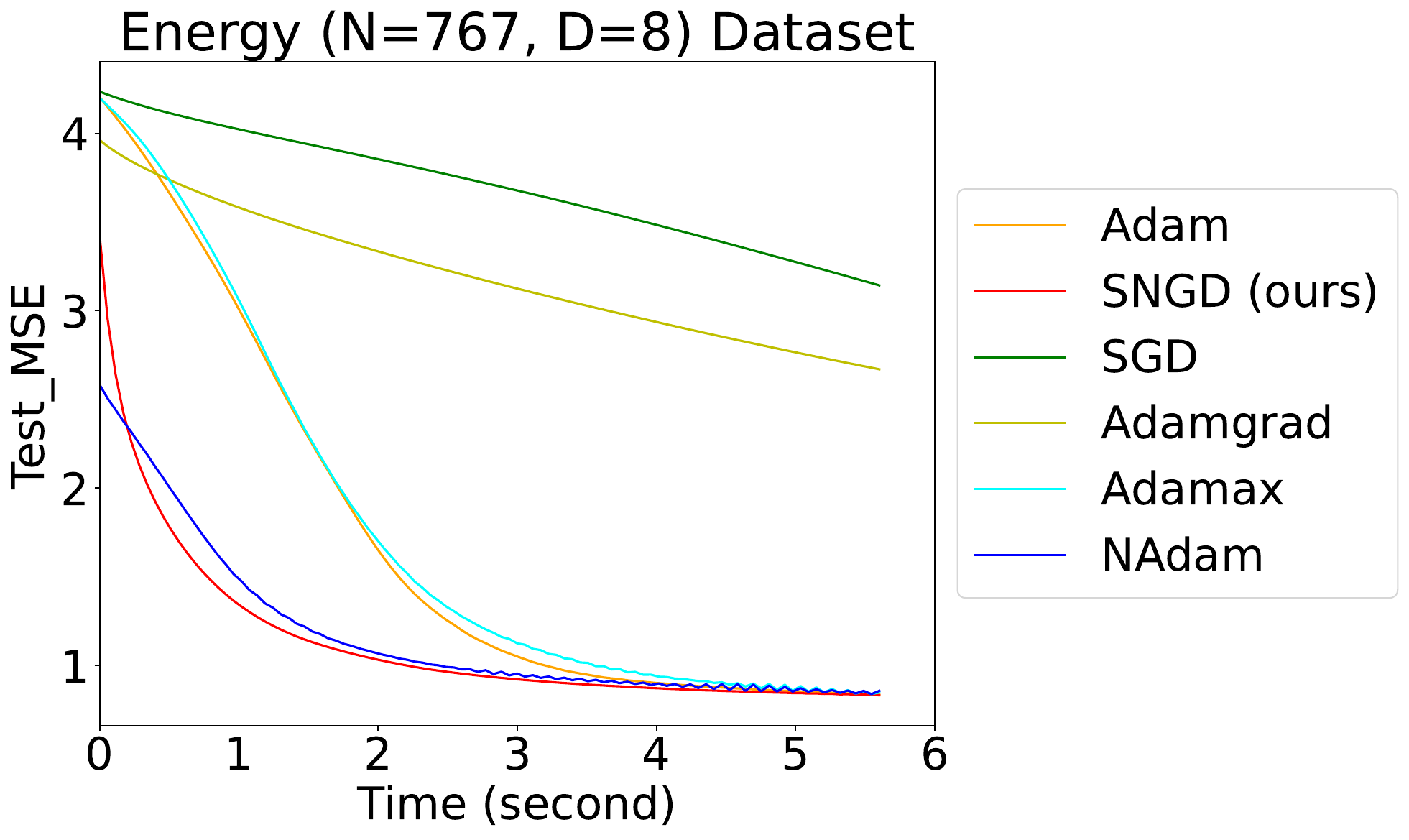}\quad
        \includegraphics[width=.49\textwidth]{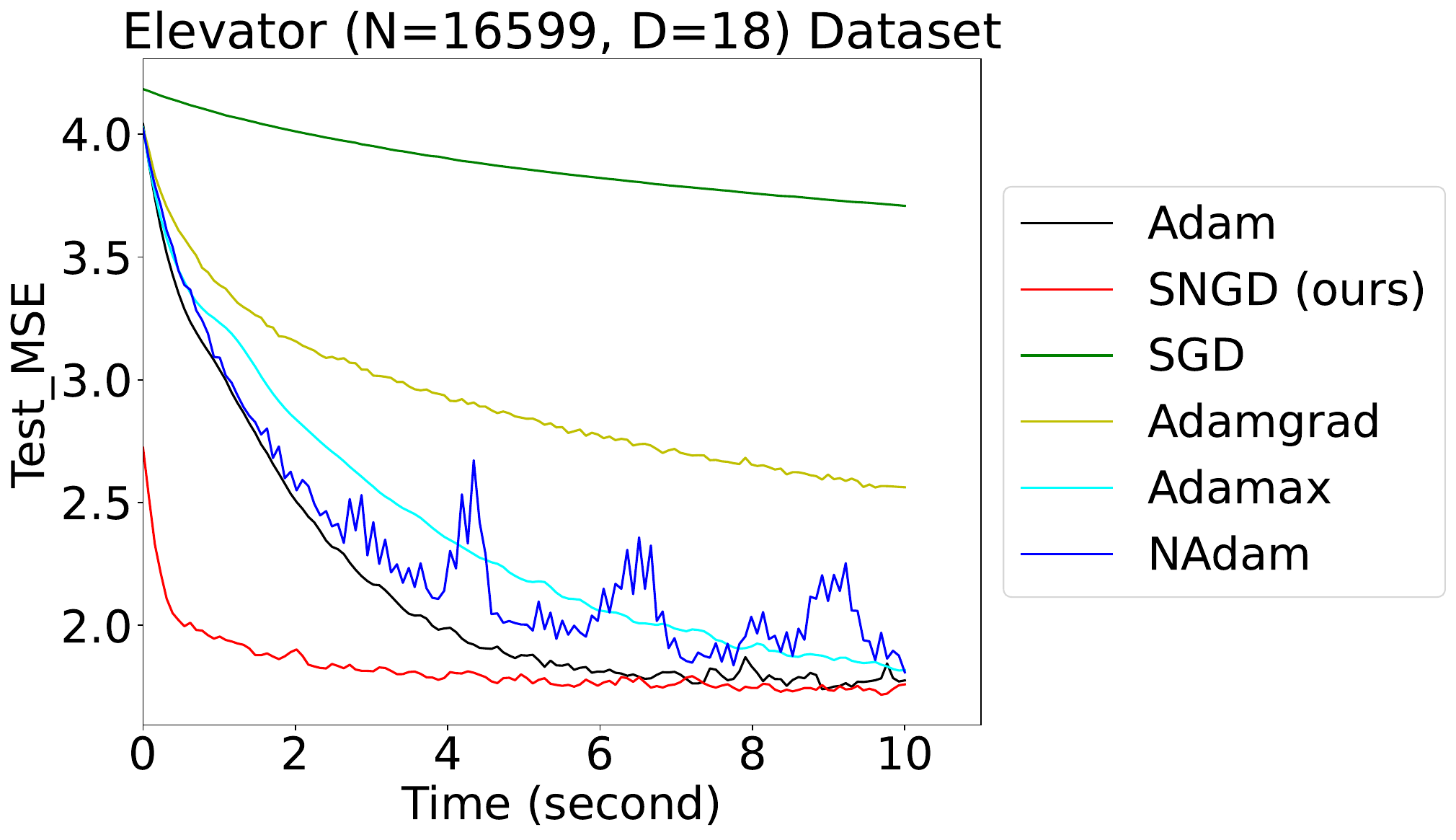}
        \includegraphics[width=.49\textwidth]{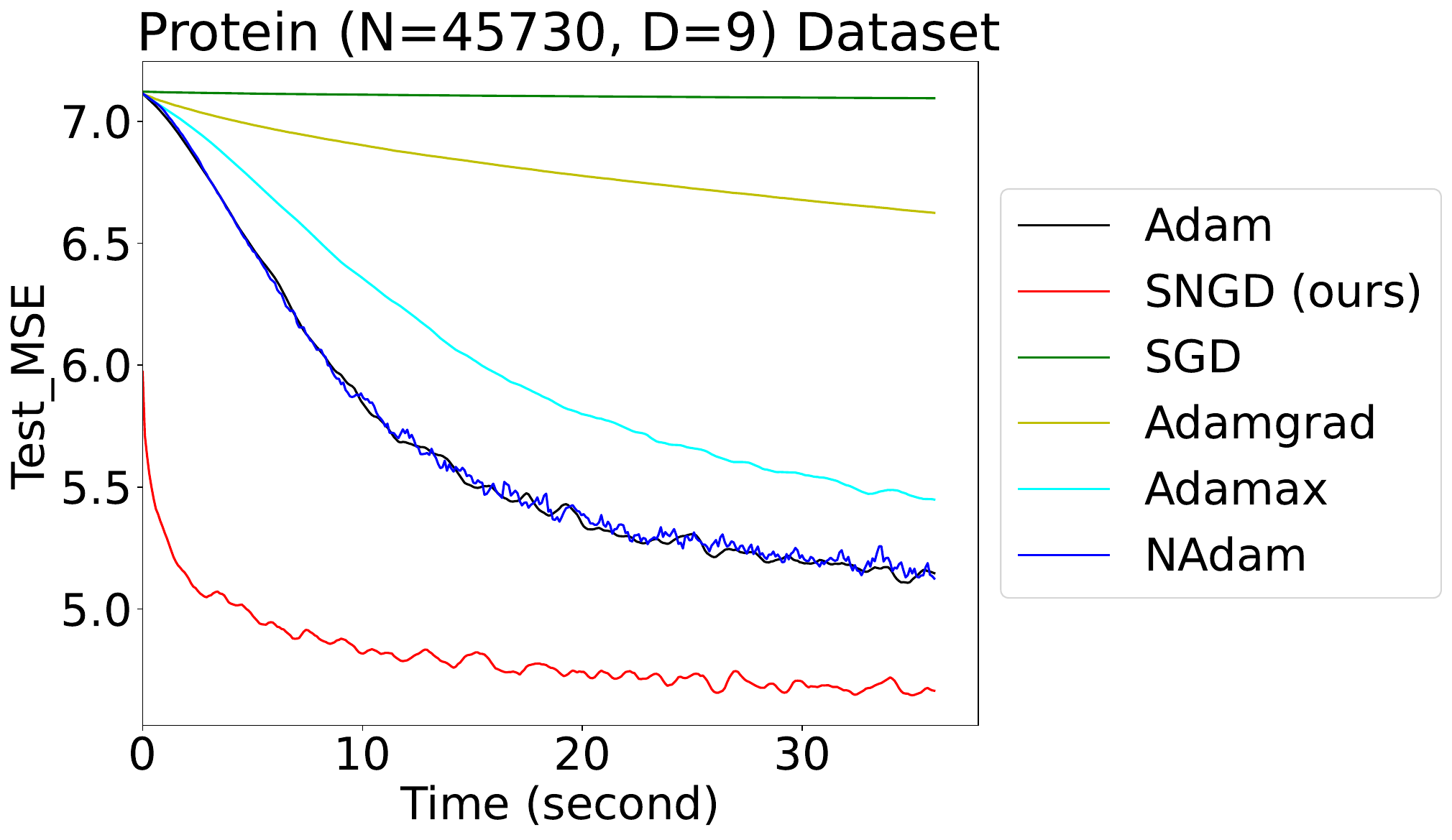}\quad
        \includegraphics[width=.49\textwidth]{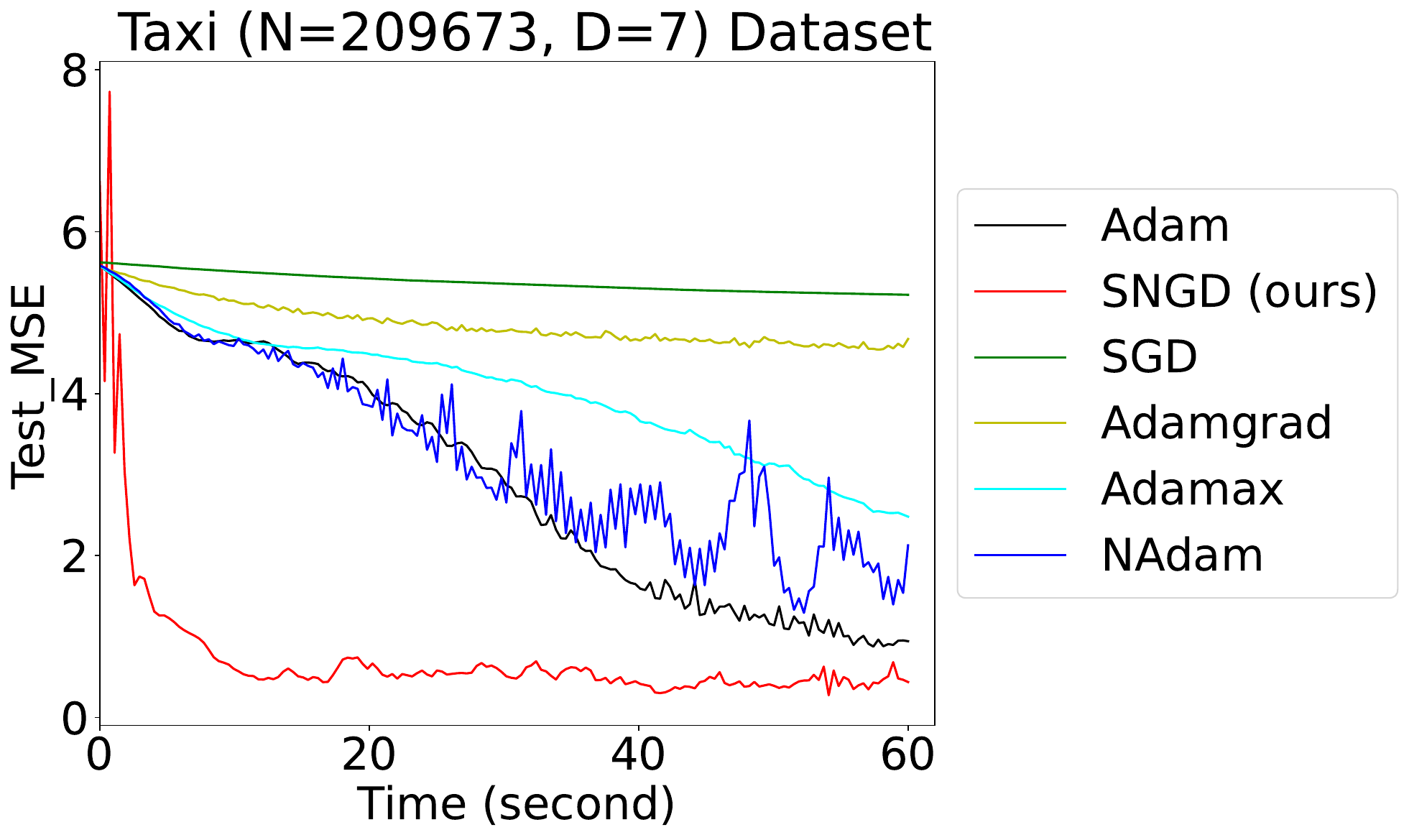}
\caption{Test MSE curves for four datasets. Our SNGD method (red) demonstrates faster convergence and lower final test error compared to five baseline optimizers (Adam, SGD, Adagrad, Adamax, and Nadam).}
\label{fig:main1}
\end{figure*}
\subsection{SNGD Algorithm for SVTP}
We further evaluate the efficiency of the proposed SNGD algorithm (Algorithm~\ref{algorithm}). The training negative ELBO curves for four datasets are shown in Figure~\ref{fig:main}. 

\textbf{Convergence Analysis:} Our method achieves the fastest convergence across Energy, Protein, and Taxi datasets. Specifically, on the Protein dataset, SNGD shows dramatic early improvement, reaching near-optimal ELBO values within the first 15 seconds. The method remains competitive on Elevator, where Adam initially appears faster but SNGD achieves better final convergence after 4 seconds. The pronounced fluctuations observed on the Taxi dataset are likely attributable to extreme outliers that cause temporary increases in the loss function.

\textbf{Theoretical Explanation:} The improved convergence of SNGD stems from its geometry-aware updates. Unlike standard gradient methods that treat all directions in the parameter space equally, natural gradients account for the underlying geometry by preconditioning with the inverse Fisher information matrix. This leads to more direct optimization paths that align with the true structure of the variational distribution family, resulting in faster and more stable convergence.

\textbf{Generalization Performance:} To assess generalization, we evaluate all algorithms under identical conditions on the testing set using MSE as the metric, as shown in Figure~\ref{fig:main1}. Our method consistently attains the best (lowest) final test error across all four datasets. On Energy, SNGD converges to MSE $\approx 0.72$ while Adam reaches only $\approx 0.75$. Similar advantages are observed on Elevator (SNGD: 1.83 vs Adam: 1.85) and Protein (SNGD: 4.84 vs Adam: 4.86).

\textbf{Comparison with Baseline Optimizers:} The relationship between training convergence (Figure~\ref{fig:main}) and test performance (Figure~\ref{fig:main1}) reveals useful insights. While SGD is stable, it converges too slowly for practical use, requiring significantly more iterations to reach comparable performance. Adam, despite being a strong general-purpose optimizer with adaptive learning rates, tends to get trapped in suboptimal regions of the loss. In contrast, SNGD's geometry-aware optimization leads to more accurate posterior estimates, which directly translates to improved generalization performance. This advantage is particularly pronounced on datasets with complex distributions (Protein, Taxi), where capturing the correct posterior geometry is critical for robust predictions.

\textbf{Comparison with Recent Robust Methods:} We further compare SNGD-SVTP with recent robust baseline methods: Robust SVGPR (RSVGPR) \cite{huang2024robust} and NOVI \cite{xu2024neural}. RSVGPR extends SVGP with robust likelihood approximations, while NOVI employs neural variational inference with implicit distributions. Table~\ref{tab:robust_comparison} presents the test MSE comparison across four datasets. As shown, RSVGPR improves upon standard SVGP by incorporating robust likelihood modeling, while NOVI, with its neural variational inference and higher model capacity, achieves strong performance on datasets with complex patterns. Nevertheless, our SNGD-SVTP method consistently outperforms both baselines across all datasets. This advantage arises from: ($i$) end-to-end Student-t modeling that enhances robustness in both prior and likelihood, ($ii$) natural gradient optimization that enables efficient posterior exploration, and ($iii$) a principled sparse approximation that balances computational efficiency and model expressiveness.

\renewcommand{\arraystretch}{1.1}{
\begin{table}[t]
\centering
\resizebox{1 \columnwidth}{!}{
\begin{tabular}{lcccc}
\hline
\textbf{Method} & \textbf{Energy} & \textbf{Elevator} & \textbf{Protein} & \textbf{Taxi} \\
\hline
SVGP & 2.02 & 6.54 & 7.23 & 0.76 \\
RSVGPR \cite{huang2024robust} & 1.82 & 5.89 & 6.51 & 0.68 \\
NOVI \cite{xu2024neural} & 0.85 & 2.15 & 5.20 & 0.48 \\
SNGD-SVTP (Ours) & \textbf{0.72} & \textbf{1.83} & \textbf{4.84} & \textbf{0.41} \\
\hline
\end{tabular}
}
\caption{A comparison of test MSE with other recent robust methods, demonstrating improved performance.}
\label{tab:robust_comparison}
\end{table}
}

\section{Conclusion}
\label{sec.con}

We have proposed a novel framework for Sparse Variational Student-t Processes (SVTP) that addresses the computational challenges inherent in modeling heavy-tailed distributions and the need for robustness to outliers. By leveraging conditional representations and sparse inducing points, the proposed framework enhances the flexibility and scalability of Student-t processes for large-scale applications. Furthermore, by incorporating natural gradient methods from information geometry theory, we improved the optimization of variational parameters compared to conventional gradient-based optimizers such as Adam. The integration of the Fisher information matrix within a mini-batch stochastic framework further enhances the computational efficiency of natural gradients.

Experiments on synthetic data and real-world datasets from UCI and Kaggle demonstrate that the proposed methods consistently outperform baseline approaches in terms of convergence speed, computational efficiency, predictive accuracy, and robustness to outliers. These empirical findings highlight the effectiveness of SVTP in handling non-Gaussian noise and outlier-dominated environments.

Overall, this work contributes to the growing literature on non-Gaussian process nonparametric models for function approximation and provides a scalable and robust alternative to Gaussian-based approaches. Future research directions include extending the SVTP framework to more complex settings, such as temporal models, high-dimensional data, and structured prediction tasks.

\section*{Acknowledgement}
This work was supported in part by grants from National Natural Science Foundation of China 
(52539005), the fundamental research program of Guangdong, China (2023A1515011281), the China Scholarship Council (202306150167), Guangdong Basic and Applied Basic Research Foundation (24202107190000687), Foshan Science and Technology Research Project (2220001018608).


\numberwithin{lemma}{section} 
\renewcommand{\thelemma}{\thesection.\arabic{lemma}}
\numberwithin{theorem}{section}
\renewcommand{\thetheorem}{\thesection.\arabic{theorem}}

\renewcommand{\theproposition}{\thesection.\arabic{proposition}}
\appendix

\subsection{Proof of Lemmas and Theorems.}
\begin{lemma}
\label{l1a}
Suppose that $y \sim \mathcal{ST}_n(\nu,\phi,K)$ is partitioned into $y_1$ and $y_2$ with dimensions $n_1$ and $n_2$, respectively. Let $\phi_1$, $\phi_2$, $K_{11}$, $K_{12}$, and $K_{22}$ denote the corresponding partitioned matrices,
$$\phi = (\phi_1, \phi_2), \qquad K = \begin{pmatrix}K_{11} & K_{12} \\ K_{12}^\top  & K_{22}\end{pmatrix}.$$ 
Then, given $y_1$, the conditional distribution of $y_2$ can be expressed as $\mathcal{ST}_{n_2}(\nu+n_1,\phi_2^*,\frac{\nu+\beta_1-2}{\nu+n_1-2}k^*)$, where $\phi_2^* = K_{21}K_{11}^{-1}(y_1 - \phi_1) + \phi_2, \beta_1 = (y_1 - \phi_1)^\top K_{11}^{-1}(y_1 - \phi_1)$ and $k^* = K_{22} - K_{21}K_{11}^{-1}K_{12}$. The mean and covariance matrix of the conditional distribution of $y_2$ given $y_1$ are $\mathbb{E}[y_2|y_1] = \phi_2^*$ and $\text{cov}[y_2|y_1] = \frac{\nu+\beta_1-2}{\nu+n_1-2}k^*$, respectively.
\end{lemma}
\begin{proof}
 Let $\beta_2=\left(\boldsymbol{y}_{\mathbf{2}}-\boldsymbol{\phi}^*_2\right)^{\top} t^{*-1}\left(\boldsymbol{y}_2-\boldsymbol{\phi}^*_2\right)$ and note that $\beta_1+\beta_2=(\boldsymbol{y}-\boldsymbol{\phi})^{\top} K^{-1}(\boldsymbol{y}-\boldsymbol{\phi})$. We have
\begin{equation*}
\begin{aligned}\textstyle
p\left(\boldsymbol{y}_{\mathbf{2}} \mid \boldsymbol{y}_{\mathbf{1}}\right)&=\frac{p\left(\boldsymbol{y}_{\mathbf{1}}, \boldsymbol{y}_{\mathbf{2}}\right)}{p\left(\boldsymbol{y}_{\mathbf{1}}\right)} \\& \textstyle\propto\left(1+\frac{\beta_1+\beta_2}{\nu-2}\right)^{-(\nu+n) / 2}\left(1+\frac{\beta_1}{\nu-2}\right)^{\left(\nu+n_1\right) / 2} \\
&\textstyle \propto\left(1+\frac{\beta_2}{\beta_1+\nu-2}\right)^{-(\nu+n) / 2}
\end{aligned}
\end{equation*}
Comparing this expression to the definition of a $\mathcal{ST}$ density function gives the required result.
\end{proof}

\begin{lemma}
\label{ls}
Let $K_n$ be an $n \times n$ symmetric positive definite matrix, $\phi \in \mathbb{R}^n$, $\nu >0$ and $\rho >0$. If
\begin{equation}\nonumber
r^{-1} \sim \Gamma(\nu / 2, \rho / 2),~~~
\boldsymbol{y} \mid r  \sim \mathcal{N}(\boldsymbol{\phi}, r(\nu-2) K_n / \rho),
\end{equation}
then marginally $
\boldsymbol{y}\sim \mathcal{S} \mathcal{T} (\nu,\phi,K_n)$.   \end{lemma}
\begin{proof}
     Let $\beta=(\boldsymbol{y}-\boldsymbol{\phi})^{\top} K^{-1}(\boldsymbol{y}-\boldsymbol{\phi})$. We can analytically marginalize out the scalar $r$,
$$
\begin{aligned}
p(\boldsymbol{y}) &\textstyle=\int p(\boldsymbol{y} \mid r) p(r) d r\\ & \textstyle\propto \int \exp \left(-\frac{\rho \beta}{2(\nu-2) r}\right) r^{-\frac{n}{2}} \exp \left(-\frac{\rho}{2 r}\right) r^{-\frac{(\nu+2)}{2}} d r \\
&\textstyle \propto\left(1+\frac{\beta}{\nu-2}\right)^{-\frac{(\nu+n)}{2}} \int \exp \left(-\frac{1}{2 r}\right) r^{-\frac{(\nu+n+2)}{2}} d r \\
\end{aligned}
$$
The integral equals 1, hence $\boldsymbol{y} \sim \mathcal{ST}(\nu, \boldsymbol{\phi}, K)$.
\end{proof}

\begin{lemma}
Let $X$ be an $n$-dimensional Gaussian random vector with mean $\mu$ and covariance $\Sigma$. Then for $A \in \mathbb{R}^{n\times n}$, $b \in \mathbb{R}^n$, the random
vector $Y = AX + b$ is distributed as $\mathcal{N}(A\mu + b,A^\top \Sigma A)$.
\end{lemma}

\begin{proof}
 Let $\phi_X(t)=\mathbb{E}\left[\exp \left(i t^{\top} X\right)\right]$ be the characteristic function of a random variable $X \in \mathbb{R}^n$.
If $X \sim \mathcal{N}(\mu, \Sigma)$, then $\phi_X(t)=\exp \left(i t^{\top} \mu-\frac{1}{2} t^{\top} \Sigma t\right)$
If $Y = AX + b$, then a straightforward sequence can show that
$$
\begin{aligned}
\phi_Y(t) & =\mathbb{E}\left[\exp \left(i t^{\top}(A X+b)\right)\right] \\
& =\exp \big(i t^{\top}(A \mu+b)-\tfrac{1}{2} t^{\top} A \Sigma A^{\top} t\big)
\end{aligned}
$$
Since the characteristic function uniquely defines the distribution, we have $Y \sim \mathcal{N}\left(A \mu+b, A \Sigma A^{\top}\right)$.   
\end{proof}

\begin{lemma}
\label{l3}
     Let $ p_X(\boldsymbol{x})=|\Sigma|^
{-\frac{1}{2}}p_{X_0}( \Sigma^{-\frac{1}{2}}
(\boldsymbol{x}-\mu) )$ be a location-scale probability density function,
where $\mu \in \mathbb{R}^n$ is the location vector and $\Sigma \in  \mathbb{R}^ {n\times n}$ is the dispersion matrix. Let $X_0 =
\Sigma^{-\frac{1}{2}}
(X-\mu )$ be a standardized version of $X$ that does not depend on $(\mu,\Sigma )$. Then
\begin{equation}
    \mathbb{E}_{p(X)}[f(\Sigma^{-\frac{1}{2}}
(X-\mu ))]=\mathbb{E}_{p(X_0)}[f(X_0)]
\end{equation}
where $f$  is any arbitrary continuous function.
\end{lemma}
\begin{proof} Apply a simple change of variables.
\end{proof}

\begin{lemma}

let $X_0$ be an N-dimensional standard student-t
random vector, $p(X_0)=\mathcal{ST}(\nu,0,I)$, then
\begin{equation}\nonumber
\textstyle   \mathbb{E} _{p\left( X_0 \right)}\left[ \log \left( 1+\frac{{X_0}^{\top}X_0}{\nu-2} \right) \right] =\Psi \left( \frac{\nu +N}{2} \right) -\Psi \left( \frac{\nu}{2} \right) 
\end{equation}
where $\Psi(\cdot) $  is the digamma function. 
\end{lemma}
\begin{proof}
    See, e.g., \cite{kotz2004multivariate}.
\end{proof}

\begin{theorem}
 An upper bound for $\mathcal{L}_2$ is
 \begin{small}
     \begin{equation*}
 \label{l2a}
     \mathcal{L}_2 \leq
\log \left\{ 1+\frac{1}{\nu-2}\mathrm{Tr}\left( K_{Z,Z'}^{-1}\mathbf{S} \right) +\frac{1}{\nu -2}\mathrm{Tr}\left( K_{Z,Z'}^{-1}\mathbf{mm}^\top  \right) \right\} 
\end{equation*}
\end{small}
\end{theorem}
\begin{proof} Let $\mathbf{S} = \mathbb{E}[(\mathbf{u}-\mathbf{m})(\mathbf{u}-\mathbf{m})^\top]$ and use the simple equality $\mathbf{u}= \mathbf{u}-\mathbf{m}+\mathbf{m}$. Then
\begin{small}
    \begin{equation*}
\begin{aligned}
	\mathcal{L}_2=~&\textstyle\mathbb{E} _{q(\mathbf{u})}\left\{ \log \left[ 1+\frac{1}{\nu -2}\mathbf{u}^{\mathrm{T}}K_{Z,Z'}^{-1}\mathbf{u} \right] \right\}\\
	\le ~&\textstyle\log \left\{ \mathbb{E} _{q(\mathbf{u})}\left[ 1+\frac{1}{\nu -2}\mathbf{u}^{\mathrm{T}}K_{Z,Z'}^{-1}\mathbf{u} \right] \right\}\\
	=~&\textstyle\log \left\{ 1+\frac{1}{\nu -2}\mathrm{tr}\left( K_{Z,Z'}^{-1}\mathbf{S} \right) +\frac{1}{\nu -2}\mathrm{tr}\left[ K_{Z,Z'}^{-1}\mathbf{mm}^{\mathrm{T}} \right] \right\}
\end{aligned}
\end{equation*}
\end{small}
\end{proof}
\begin{lemma}
    Let  $ f \sim \mathcal{T} \mathcal{P}(\nu, \phi, K)$ and $g \sim \mathcal{G} \mathcal{P}(\phi, K)$.  Then $f \stackrel{d}{\longrightarrow} g$ as $\nu \rightarrow \infty$. 
\end{lemma}

\begin{proof}
     It suffices to show convergence for any finite set of points. Let $\boldsymbol{y} \sim \mathcal{ST}_n(\nu, \boldsymbol{\phi}, K)$ and $\beta=(\boldsymbol{y}-\boldsymbol{\phi})^{\top} K^{-1}(\boldsymbol{y}-\boldsymbol{\phi})$ then as $\nu \rightarrow \infty$,
$$\textstyle p(\boldsymbol{y}) \propto\left(1+\frac{\beta}{\nu-2}\right)^{-(\nu+n) / 2} \longrightarrow ~ e^{-\beta / 2}.$$
Therefore the distribution of $\boldsymbol{y}$ converges to $\mathcal{N}_n(\phi, K)$.
\end{proof}
\begin{theorem}
  as $\nu\rightarrow\infty$, the posterior distribution of SVTP converges in distribution to the posterior distribution of SVGP.
\end{theorem}
\begin{proof}
 For the SVTP, the joint density of $\mathbf{u}$ and $\mathbf{f}$ is
 $$
p(\mathbf{u},\mathbf{f})=\mathcal{S} \mathcal{T} \left( \nu,\left[ \begin{array}{c}
	0\\
	0\\
\end{array} \right],\left[ \begin{matrix}
	K_{Z,Z'}&		K_{Z,X}\\
	K_{X,Z}&		K_{X,X'}\\
\end{matrix} \right] \right),
$$
and as $\nu\rightarrow\infty$
$$\mathbf{f},\mathbf{u}\xrightarrow{\mathcal{P}}\mathbf{f}',\mathbf{u}'\sim \mathcal{N} \left( \left[ \begin{array}{c}
	0\\
	0\\
\end{array} \right],\left[ \begin{matrix}
	K_{Z,Z'}&		K_{Z,X}\\
	K_{X,Z}&		K_{X,X'}\\
\end{matrix} \right] \right).
$$
By Bayes Rule, the the posterior distribution of SVTP is $p( \mathbf{u},\mathbf{f}|\mathbf{y} ) =p( \mathbf{f},\mathbf{u} ) p( \mathbf{y}|\mathbf{f} )/p( \mathbf{y} )$. By continuity of the random variables, as $\nu \rightarrow \infty$,  $\mathbf{u},\mathbf{f}|\mathbf{y}  \xrightarrow{\mathcal{P}} \mathbf{u}',\mathbf{f}'|\mathbf{y}$.

\end{proof}
\begin{proposition}
When $q(\mathbf{u})=\mathcal{ST}(
\nu +n, \mathbf{m},\mathbf{S})$, the marginal distribution of $\mathbf{u}$ is
$q( \mathbf{f} ) =\int{p( \mathbf{f}|\mathbf{u} ) q( \mathbf{u} )}d\mathbf{u}=\mathcal{ST}\left( \nu,\mu ',\Sigma ' \right)$,
where the mean $\mu'=K_{X,Z}K_{Z,Z'}^{-1}\mathbf{m}$ and the covariance matrix $\Sigma'=K_{X,X'}-K_{X,Z}K_{Z,Z'}^{-1}(K_{Z,Z'}-\mathbf{S})K_{Z,Z'}^{-1}K_{Z,X}$.
\end{proposition}
\begin{proof}
Define $\beta _1=( \mathbf{u}-\mathbf{m} ) ^{\top}\mathbf{S}^{-1}( \mathbf{u}-\mathbf{m} )$, and
$\beta _2=( \mathbf{f}-\mu ) ^{\top}\Sigma ^{-1}( \mathbf{f}-\mu  ) $, where $\mu=K_{X,Z}K_{Z,Z'}^{-1}\mathbf{u}$ and $\Sigma=K_{X,X'}-K_{X,Z}K_{Z,Z'}^{-1}K_{Z,X}$. Then
\begin{equation*}
\begin{aligned}
     p( \mathbf{f}|\mathbf{u} ) q( \mathbf{u} )  \propto&\textstyle \left( 1+\frac{\beta _2}{\beta _1+\nu -2} \right) ^{-\frac{\nu +n+M}{2}}\left( 1+\frac{\beta _1}{\nu -2} \right) ^{-\frac{\nu +n+M }{2}}\\
    \propto&\textstyle \left( 1+\frac{\beta _1+\beta _2}{\nu -2} \right) ^{-\frac{\nu +n+M}{2}}.
\end{aligned} 
\end{equation*}
This is a multivariate Student's t-distribution. We have
\begin{equation*}
\begin{aligned}
    \beta _1+\beta _2=\left[ \begin{matrix}
	\mathbf{u}-\mathbf{m}&		\mathbf{f}-\mu '\\
\end{matrix} \right] \left[ \begin{matrix}
	\mathbf{S}&		\Sigma _{12}\\
	\Sigma _{21}&		\Sigma '\\
\end{matrix} \right] ^{-1}\left[ \begin{array}{c}
	\mathbf{u}-\mathbf{m}\\
	\mathbf{f}-\mu '\\
\end{array} \right] 
\\
\end{aligned}
\end{equation*}
where $\mu'$ and $\Sigma'$ are as given and $\Sigma_{21}=\Sigma_{12}^\top =K_{X,Z}K_{Z,Z'}^{-1}\mathbf{S}$ is the covariance of $\mathbf{u}$ and $\mathbf{f}$. From this, we obtain the marginal distribution $q( \mathbf{f} )=\mathcal{ST}( \nu,\mu ',\Sigma ' )$.
\end{proof}

\subsection{A Detailed Proof of Proposition \ref{prop:beta_link}}
\label{appendix:beta_link}

Let
$\mathbf{x} \sim t_{\tilde{\nu}}\!\left(\mathbf{m},\mathbf{S}\right)$ and $\mathbf{S}=\mathrm{diag}(\sigma_1^{2},\ldots,\sigma_M^{2})$, and define $\boldsymbol{\xi}
=\mathbf{S}^{-1/2}(\mathbf{x}-\mathbf{m})$. For $i\neq j$, the function $\xi_i(1+\|\boldsymbol{\xi}\|^{2})^{-a}$ is odd in $\xi_i$. Therefore
\begin{equation}\nonumber\textstyle
\int_{\mathbb{R}} \xi_i(1+\|\boldsymbol{\xi}\|^{2})^{-a} d\xi_i=0,
\end{equation}
and by Fubini $F_{ij}^{\mathbf{m}}=0$, $F^{\mathbf{m}\mathbf{S}}=\mathbf{0}$, and $F^{\mathbf{m}\tilde{\nu}}=\mathbf{0}$. For $i=j$,
\begin{equation}\nonumber\textstyle
\int_{\mathbb{R}^{M}}
\xi_i^2 g(\|\boldsymbol{\xi}\|^2)\, d\boldsymbol{\xi}
=
\frac{1}{M}
\int_{\mathbb{R}^{M}}
\|\boldsymbol{\xi}\|^2 g(\|\boldsymbol{\xi}\|^2)\, d\boldsymbol{\xi}.
\end{equation}
Let $r=\|\boldsymbol{\xi}\|$. The Jacobian satisfies
\begin{equation}\nonumber\textstyle
d\boldsymbol{\xi}
=
\big( r^{M-1}dr\big)d\eta_1\cdots d\eta_{M-1}\prod_{m=1}^{M-2} \sin^{M-m-1} \eta_m.
\end{equation}
The integral can be solved using Wallis’ formula \cite{kazarinoff1956wallis,guo2015wallis},
\begin{equation}\nonumber\textstyle
\int_{0}^{\pi/2} (\sin \eta)^{p}\, d\eta
=
\frac{\sqrt{\pi}}{2}
\frac{\Gamma\!\left(\frac{p+1}{2}\right)}
     {\Gamma\!\left(\frac{p}{2}+1\right)}
\end{equation}
for $p\in\mathbb{N}$. Applying this recursively yields
\begin{equation}\nonumber\textstyle
\int d\eta_1\cdots d\eta_{M-1}\prod_{m=1}^{M-2} \sin^{M-m-1} \eta_m =\frac{2\pi^{M/2}}{\Gamma(M/2)}.
\end{equation}
Thus,
\begin{equation}\nonumber\textstyle
\int_{\mathbb{R}^{M}}
\frac{\|\boldsymbol{\xi}\|^{2}}{(1+\|\boldsymbol{\xi}\|^{2})^{a}}
\, d\boldsymbol{\xi}
=
\frac{\pi^{\frac{M}{2}}}{\Gamma(\frac{M}{2})}
\int_{0}^{\infty}
r^{M+1}(1+r^{2})^{-a}\, dr.
\end{equation}
Let $t=r^{2}$. Then
\begin{equation}\nonumber\textstyle
\int_{0}^{\infty}
r^{M+1}(1+r^{2})^{-a} dr
=
\frac{1}{2}
\int_{0}^{\infty}
t^{(M+3)/2 -1}(1+t)^{-a}\,dt.
\end{equation}
Using the beta function $\mathrm{B}(x,y)
=\int_{0}^{\infty} t^{x-1}(1+t)^{-x-y}\,dt$ we see that $x = \frac{M+3}{2}$ and $y = a - \frac{M+3}{2}$. Thus,
\begin{equation}\nonumber\textstyle
\textstyle \int_{\mathbb{R}^{M}}
\xi_i^{2}
(1+\|\boldsymbol{\xi}\|^{2})^{-a}\, d\boldsymbol{\xi}
=
\frac{1}{M}
\frac{\pi^{M/2}}{\Gamma(M/2)}
\mathrm{B}\!\left(\frac{M+3}{2},\, a-\frac{M+3}{2}\right).
\end{equation}
Setting $a=\frac{\tilde{\nu}+M}{2}+2$ gives
\begin{equation}\nonumber\textstyle
\int_{\mathbb{R}^{M}}
\xi_i^{2}
(1+\|\boldsymbol{\xi}\|^{2})^{-(\tilde{\nu}+M)/2 -2}\,
d\boldsymbol{\xi}
=
\frac{1}{M}
\frac{\pi^{M/2}}{\Gamma(M/2)}
\mathrm{B}\!\left(
\frac{M+3}{2},\,\frac{\tilde{\nu}+1}{2}
\right).
\end{equation}
Substituting this expression into the Fisher Information components for
$\mathbf{m}$, $\tilde{\nu}$, $\mathbf{S}$, and the cross terms yields the block expressions for $F^{\mathbf{m}}$, $F^{\tilde{\nu}}$, $F^{\mathbf{S}}$, and $F^{\tilde{\nu}\mathbf{S}}$, given in Proposition~\ref{prop:beta_link}.  All mixed blocks involving $\mathbf{m}$ disappear due to symmetry.
\hfill$\square$

\vskip 0.2in

\bibliography{sample}
\bibliographystyle{IEEEtran}

\end{document}